\documentclass[pdflatex,sn-mathphys-ay]{sn-jnl}% Math and Physical Sciences Numbered Reference Style
\usepackage{graphicx}%
\usepackage{multirow}%
\usepackage{amsmath,amssymb,amsfonts}%
\usepackage{amsthm}%
\usepackage{mathrsfs}%
\usepackage[title]{appendix}%
\usepackage{xcolor}%
\usepackage{textcomp}%
\usepackage{manyfoot}%
\usepackage{booktabs}%
\usepackage{algorithm}%
\usepackage{algorithmicx}%
\usepackage{algpseudocode}%
\usepackage{listings}%
\usepackage{colortbl}
\usepackage{dsfont}
\usepackage{multicol} 
\usepackage{pifont}
\usepackage{subcaption}
\usepackage{tikz}
\usepackage{pgf}
\usepackage{url}
\usepackage{calc}
\usepackage{soul}

\newcommand{\quantifier}{q}
\newcommand{\Ptr}{P_{\text{tr}}}
\newcommand{\Pte}{P_{\text{te}}}
\newcommand{\sample}{S}

\newcommand{\FP}{\mathsf{FP}}
\newcommand{\FN}{\mathsf{FN}}
\newcommand{\tpr}{\operatorname{\textsf{tpr}}}
\newcommand{\fpr}{\operatorname{\textsf{fpr}}}

\newcommand{\killpunct}[1]{}

\begin{document}

\title{Efficient Quantification on Large-Scale Networks}

%% =============================================================%%
%% GivenName -> \fnm{Joergen W.} Particle -> \spfx{van der} -> surname
%% prefix FamilyName -> \sur{Ploeg} Suffix -> \sfx{IV}
%% \sfx{IV}}\email{iauthor@gmail.com}
%% =============================================================%%

\author[1]{\fnm{Alessio}
\sur{Micheli}}\email{alessio.micheli@unipi.it} \equalcont{These
authors contributed equally to this work.}

\author[2]{\fnm{Alejandro}
\sur{Moreo}}\email{alejandro.moreo@isti.cnr.it} \equalcont{These
authors contributed equally to this work.}

\author[1]{\fnm{Marco} \sur{Podda}}\email{marco.podda@unipi.it}
\equalcont{These authors contributed equally to this work.}

\author[2]{\fnm{Fabrizio}
\sur{Sebastiani}}\email{fabrizio.sebastiani@isti.cnr.it}
\equalcont{These authors contributed equally to this work.}

\author[]{\fnm{William} \sur{Simoni}}\email{wilsimoni@gmail.com}
\equalcont{These authors contributed equally to this work.}

\author*[1]{\fnm{Domenico}
\sur{Tortorella}}\email{domenico.tortorella@unipi.it}
\equalcont{These authors contributed equally to this work.}

\affil*[1]{\orgdiv{Dipartimento di Informatica}, \orgname{Università
di Pisa}, \orgaddress{\street{Largo Bruno Pontecorvo, 3}, \city{Pisa},
\postcode{56127}, \country{Italy}}}

\affil[2]{\orgdiv{Istituto di Scienza e Tecnologie dell’Informazione},
\orgname{Consiglio Nazionale delle Ricerche}, \orgaddress{\street{Via
Giuseppe Moruzzi 1}, \city{Pisa}, \postcode{56124}, \country{Italy}}}


\abstract{\textit{Network quantification} (NQ) is the problem of
estimating the proportions of nodes belonging to each class in subsets
of unlabelled graph nodes. When prior probability shift is at play,
this task cannot be effectively addressed by first classifying the
nodes and then counting the class predictions. In addition, unlike
non-relational quantification, NQ demands
enhanced flexibility in order to capture a broad range of connectivity
patterns, resilience to the challenge of heterophily, and scalability
to large networks. In order to meet these stringent requirements, we
introduce XNQ, a novel method that synergizes the flexibility and
efficiency of the unsupervised node embeddings computed by randomized
recursive Graph Neural Networks, with an Expectation-Maximization
algorithm that provides a robust quantification-aware adjustment to
the output probabilities of a calibrated node classifier. 
In an extensive evaluation, in which we also validate
the design choices underpinning XNQ through comprehensive
ablation experiments, we find that XNQ
consistently and significantly improves on the best network
quantification methods to date, thereby setting the new state of the
art for this challenging task. XNQ also provides a training
speed-up of up to 10x -- 100x over other methods based on graph
learning.}

\keywords{Quantification, Network quantification, Graph neural
networks, Graph learning, Reservoir computing}

\maketitle

% -----------------------------------------------------------

\section{Introduction}
\label{sec:intro}
\noindent \textit{Quantification}~\citep{Esuli:2023os,
Gonzalez:2017it} is the machine learning task of estimating the
prevalence (or proportions) of each class in a set of unlabelled 
datapoints. Unlike
standard classification, which focuses on predicting a label for each
individual example, quantification works at the aggregate level by
estimating the overall fraction of unlabelled datapoints belonging to
each class. Quantification finds application in all disciplines interested in characterizing populations (such as the social sciences, political science, epidemiology, ecological modelling, and market research), but also in downstream tasks such as improving the accuracy of classifiers~\citep{Saerens:2002uq}, measuring the fairness of classifiers~\citep{Fabris:2023pj} and rankers~\citep{Jaenich:2025rr}, predicting the accuracy of classifiers~\citep{Volpi:2025bx, Volpi:2025zm}, and calibrating classifiers~\citep{Moreo:2025vn}.

Quantification methods are explicitly designed to account for
\textit{dataset shift}, which occurs when the statistical properties
of the training data differ from those of the test data, due to
changes in input features, labels, or their relationships. Most
quantification methods are tailored to one specific type of dataset
shift, namely, \emph{prior probability shift} (PPS)~\citep{Storkey:2009lp}, also referred to
as ``label shift''~\citep{Lipton:2018fj}. Specifically, 
PPS occurs when the class-conditional distribution of the input
features does not change across the training and the test data (i.e., $\Ptr(X|Y)=\Pte(X|Y)$), while
the prior distribution of the class labels does (i.e., $\Ptr(Y)\neq \Pte(Y)$).
In simple words, the
class proportions in the training set might differ significantly from
those of the test set. In data affected by PPS, standard classifiers
have been repeatedly shown to be inaccurate quantifiers, leading to
class prevalence estimates biased towards the class distribution of
the training set~\citep{Gonzalez:2024cs}. Therefore, quantification
has evolved from traditional classification, with its own models and
custom evaluation protocols that assess performances while varying the
class proportions in the test data to simulate PPS (see
Section~\ref{sec:quantification}).

This work deals with the task of \textit{network quantification}
(\textbf{NQ}), which consists of performing quantification on
interlinked datapoints, i.e., on nodes belonging to a graph. As noted
by~\citet{Tang:2010uq}, NQ is suitable in settings where the goal is
to estimate how a population of interrelated individuals is
distributed according to classes of interest (e.g., to infer the
proportion of spam or malicious accounts in a social network).

\begin{figure}[t]
 \centering
 \begin{minipage}[b][][b]{.49\textwidth}
 \centering \includegraphics[height=1.8cm]{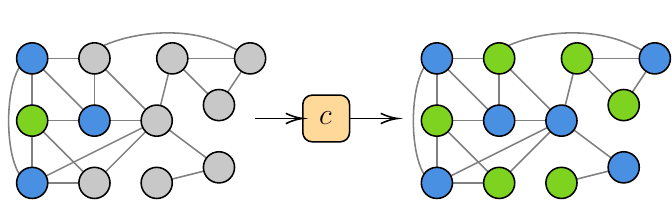}
 \subcaption{Node classification}
 \label{fig:nc}
 \end{minipage}%
 \begin{minipage}[b][][b]{.49\textwidth}
 \centering \includegraphics[height=1.8cm]{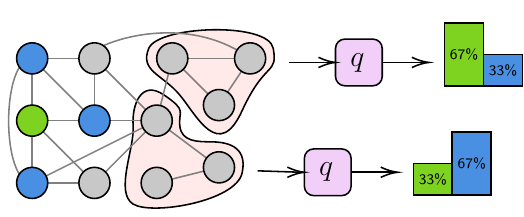}
 \subcaption{Network quantification (this work)}
 \label{fig:nq}
 \end{minipage}
 \caption{Differences between node classification and network
 quantification on a partially unlabelled graph where nodes may belong
 to the ``blue'' or ``green'' classes. Unlabelled nodes are shown in
 gray. Node classification (a) is performed by a node classifier $c$,
 which takes as input the partially unlabelled graph and returns as
 output an isomorphic graph where the class of the unlabelled nodes
 has been predicted. In contrast, network quantification (b) uses a
 quantifier $\quantifier$, which takes as input subsets of unlabelled nodes (in
 light pink shades) and returns as output their class
 distribution. In this work, we study network quantification under
 prior probability shift.}
 \label{fig:nc-vs-nq}
\end{figure}

Analogously to the distinction between quantification and
classification in non-relational domains, there is substantial
difference between NQ and the task of node classification
(Section~\ref{sec:graph-learning}). The latter is concerned with
predicting labels or assigning categories to nodes in a graph based on
their features and the network topology~\citep{bacciu2023gentle},
i.e., focusing on predicting the individual classes of the unlabelled
nodes, while the purpose of NQ is to predict their aggregate class
distribution among different network sub-communities. The example in
Figure~\ref{fig:nc-vs-nq} summarizes the main differences between
these two classes of tasks.

Due to its specific setting, NQ is also fundamentally different from
plain quantification since it is applied to the nodes of a graph,
which are not independent and identically distributed (i.i.d.) like
non-relational datapoints, but rather interdependent according to the
graph structure. Moreover, real-world networks are usually large-scale
and characterized by complex properties such as non-linear
connectivity patterns and heterophily (i.e., prevalence of inter-class
edges), which non-relational quantification models are unable to
capture.

This discussion highlights that, in order to be proficient at NQ, a
learning method should (i) exploit the network connectivity to emit
coherent prevalence predictions; (ii) be powerful enough to capture
the complex properties of real-world networks; (iii) be flexible
enough to adapt to their possibly heterophilic nature; and (iv) be
efficient and resource-friendly to make operating at scale feasible.
However, existing methods for NQ (reviewed in
Section~\ref{sec:related-work}) hardly comply with all these
requirements.

To satisfy these \textit{desiderata} we propose XNQ, a model that
integrates the representational capabilities of randomized recursive
Graph Neural Networks with a powerful Expectation-Maximization
approach for quantification. XNQ is purposely designed to be
resource-efficient, scalable, and resilient to heterophily. Through a
comprehensive experimental evaluation, we show that XNQ significantly
outperforms the best methods proposed so far for NQ, setting a new
state of the art. Additionally, we validate our design through
extensive ablation studies, showing that XNQ achieves the best
trade-off in terms of performance and computational efficiency.

The paper is organized as follows. After discussing background
concepts (Section~\ref{sec:background}) and describing related work
(Section~\ref{sec:related-work}), in Section~\ref{sec:method} we move
to detail the proposed method. In Section~\ref{sec:experiments}, XNQ
is compared against the current state-of-the-art methods on several
real-world graphs, with an in-depth ablation study and efficiency
analysis. Finally, Section~\ref{sec:conclusions} draws conclusions and
points out avenues for future research.

% -----------------------------------------------------------

\section{Background and notation}
\label{sec:background}
\noindent In this section, we introduce basic concepts and notation on
quantification and graph learning necessary to understand our
contribution.

% -----------------------------------------------------------

\subsection{Quantification}
\label{sec:quantification}
\noindent Let $\mathcal{X}$ be a generic input domain and
$\mathcal{Y}$ be a discrete set of class labels. Assume a dataset of
pairs $\mathcal{D} = \{(\boldsymbol{x}, y) \,|\, \boldsymbol{x} \in
\mathcal{X}, y \in \mathcal{Y} \}$ which gives point-wise estimates of
some unknown function we wish to learn. 
For the sake of simplicity, we formalize our method for the binary case 
$\mathcal{Y}=\{\oplus,\ominus\}$, using $\oplus$ to denote the
 ``positive'' class and $\ominus$ to denote the ``negative'' class; however, our method natively works in the multi-class case as well, as we show in Section~\ref{sec:multi-class}.  We use the symbol $\sample$ to denote a \emph{sample set}, i.e., a non-empty subset of (labelled or unlabelled) elements from $\mathcal{X}$. Let $p_{\sample}(y)$ be the true prevalence of class $y$ in sample set $\sample$ (i.e., the fraction of items in $\sample$ that belong to $y$). Note that $p_{\sample}(y)$ is just a shorthand of $\Pr(Y=y \,|\, \boldsymbol{x}\in\sample)$, where $\Pr(\cdot)$ indicates probability and $Y$ is a random variable that ranges on $\mathcal{Y}$. In the binary case, since $p_{\sample}(\ominus)=1-p_{\sample}(\oplus)$ holds true, it is sufficient to estimate the prevalence of the positive class only. A \emph{binary quantifier} $\quantifier$ is a predictor of the class prevalence $p_\sample(\oplus)$ in sample $\sample$. We characterize quantifiers by the class prevalence estimates they produce, using the notation $\hat{p}_{\sample}^{\quantifier}(y)$ to indicate that $\hat{p}_{\sample}(y)$ has been computed through $\quantifier$\footnote{We omit the superscript when the NQ method is clear from the context.}. In this work, we focus on \emph{aggregative} quantification methods, i.e., methods that work on top of a trained classifier by aggregating their individual predictions.

% -----------------------------------------------------------

\subsubsection{Quantification methods}
\noindent A trivial but inaccurate method to tackle quantification is
\emph{Classify and Count} (\textbf{CC}), which works by training a
classifier, classifying all the unlabelled datapoints, counting the
datapoints assigned to each class, and normalizing the counts so that
the result is a probability distribution over the classes. Indeed, CC
has been repeatedly shown to deliver incorrect class prevalence
estimates~\citep{Bella:2010kx, Esuli:2023os, Forman:2008kx,
Gonzalez:2017it, Gonzalez-Castro:2013fk}. One of the reasons is that
classification and quantification are characterized by different loss
functions, since (using the binary case as an example) in
classification we want to minimize some proxy of $(\FP +\FN)$, where
$\FP$ (resp.\ $\FN$) stands for the number of false positives (resp.\ negatives).
In contrast, in quantification we want to minimize some proxy of
$|\FP-\FN|$. Another reason is that data are often characterized by
dataset shift, and the algorithms we routinely use to train our
classifiers assume that no dataset shift is at play, i.e., that the
training data and the test data are i.i.d.

Many quantification methods that improve on CC, especially in terms of
robustness to PPS, have been proposed in the literature (see
Appendix~\ref{app:quantification-methods} for a detailed presentation,
and ~\cite{Esuli:2023os, Gonzalez:2017it} for more exhaustive
surveys). One of the earliest is \emph{Adjusted Classify and Count}
(\textbf{ACC}), which applies to the class prevalence estimates
generated by CC a correction based on the misclassification rates of
the classifier as estimated via $k$-fold
cross-validation~\citep{Forman:2008kx}. \emph{Probabilistic Classify
and Count} (\textbf{PCC}) and \emph{Probabilistic Adjusted Classify
and Count} (\textbf{PACC}) are probabilistic variants of CC and ACC,
in which the integer counts and the misclassification rates needed to
compute CC and ACC are replaced with their soft (i.e., probabilistic) versions~\citep{Bella:2010kx}. A radically different approach is
instead embodied in \textbf{HDy}~\citep{Gonzalez-Castro:2013fk}, a
method that views quantification as the problem of minimizing the
divergence (measured in terms of the Hellinger Distance) between two
distributions of posterior probabilities, and by
\textbf{DyS}~\citep{Maletzke:2019qd}, which uses the Topsøe distance
instead of the Hellinger distance.

% ----------------------------------------------------------

\subsubsection{Evaluating quantifiers}
\label{sec:evaluation-metrics}
\noindent The standard approach to assess the performance of a
quantifier $\quantifier$ is to simulate PPS, i.e., evaluate its estimations for
considerably different degrees of divergence between $\Ptr(Y)$ and $\Pte(Y)$.
In the binary case, the widely adopted
\textit{artificial prevalence protocol} (APP)~\citep[\S
3.4.2]{Esuli:2023os} involves randomly generating (i.e., extracting from the unlabelled data) test subsets
$\sample$ that exhibit predetermined prevalence values lying on a regular
grid, e.g., $p_{\sample}(\oplus) \in \{0.00, 0.05, ..., 0.95,
1.00\}$. 
In the general multi-class case, instead, the APP is implemented by:

\begin{enumerate}
    \item randomly generating (usually, via the Kraemer algorithm -- see \citet[\S 3.4.3]{Esuli:2023os}) vectors of prevalence values that uniformly cover the probability simplex, i.e., the space of all legitimate vectors of prevalence values given by $\Delta^{n-1}=\{p_1,\ldots,p_n:p_i\geq0,\sum_{i=1}^n p_i=1\}$, with $p_i$ the prevalence of the $i$th class and $n=|\mathcal{Y}|$ the number of classes, and then

\item extracting, from the unlabelled data, test subsets
$\sample$ each complying with one of the randomly generated prevalence vectors.
\end{enumerate}

\noindent In this experimental protocol, the class-conditional distributions across the training data and the test subsets remain stationary (meaning that $\Ptr(X|Y)=\Pte(X|Y)$), whereas the class prevalence values of a test sample ($\Pte(Y)$) may be significantly different from those encountered in the training data ($\Ptr(Y)$).
The \textit{Absolute Error} (AE), defined as:
\begin{align}
 \label{eq:ae}
 \operatorname{AE}(p_{\sample},\hat{p}_{\sample}) & =\frac{1}{n}\sum_{y\in 
 \mathcal{Y}}|\hat{p}_{\sample}(y)-p_{\sample}(y)|, 
\end{align}
\noindent between the true prevalence $p_{\sample}$ and the estimated prevalence  $\hat{p}^\quantifier_{\sample}$, averaged across  all the test subsets $\sample$ sampled via the APP, provides a concise metric of quantifier  $\quantifier$'s performance. In addition to AE, we also consider the \textit{relative absolute error} (RAE), defined as:
\begin{align}
 \label{eq:rae}
 \operatorname{RAE}(p_{\sample},\hat{p}_{\sample}) & =\frac{1}{n}\sum_{y\in 
 \mathcal{Y}}\frac{|\hat{p}_{\sample}(y)-p_{\sample}(y)|}{p_{\sample}(y)+\varepsilon},
\end{align}
\noindent where $\varepsilon$ is a smoothing factor used to avoid division by zero when $p_{\sample}(y)=0$, which we set, following \cite{Forman:2008kx}, to $\varepsilon={1/(2|S|)}$.
AE and RAE are the two \textit{de facto} standard evaluation measures for binary and multi-class quantification  \cite[\S 3.1]{Esuli:2023os}, and it has been argued \citep{Sebastiani:2020qf} that AE better mirrors the needs of some applications of quantification while RAE better mirrors others.

\begin{figure}[t]
 \centering \includegraphics[height=.45\textwidth]{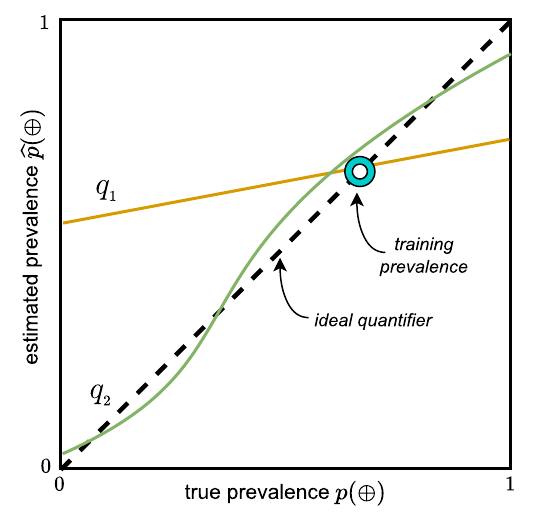}
 \caption{In the binary case, diagonal plots provide a visual tool to compare
 quantifiers. Here, $\quantifier_2$ is closer to the ideal quantifier behaviour
 (dashed diagonal), and thus superior to $\quantifier_1$.
 }
 \label{fig:diagonal-example}
\end{figure}

In the binary case, diagonal plots~\citep{Esuli:2023os} are a useful tool to visualize a
quantifier performance across different prevalence values.
Figure~\ref{fig:diagonal-example} is an example of such plots: the
$x$ axis presents the true class prevalence $p(\oplus)$, while the
$y$ axis displays the estimated prevalence $\hat{p}(\oplus)$. The dashed diagonal
represents the ideal quantifier behaviour $\hat{p}(\oplus) =
p(\oplus)$. A quantifier $\quantifier$ is visualized as a curve of points
$\left(p_{\sample}(\oplus),
 \hat{p}^\quantifier_{\sample}(\oplus)\right)$: in our example $\quantifier_2$ is
closer to the diagonal of the ideal quantifier, and thus superior to
$\quantifier_1$.

% -----------------------------------------------------------

\subsection{Graph learning}
\label{sec:graph-learning}
\noindent We define a graph $G$ by a set of nodes $\mathcal{V} = \{v_i
\,:\, 1 \leq i \leq |\mathcal{V}|\}$ and a set $\mathcal{E} \subseteq
\mathcal{V} \times \mathcal{V}$ of edges among them, encoded as an
adjacency matrix $\boldsymbol{A} \in \{0,1\}^{|\mathcal{V}| \times
|\mathcal{V}|}$ whose entries $\boldsymbol{A}_{ij}$ are $1$ if $(v_i,
v_j) \in \mathcal{E}$ and $0$ otherwise. We define the set of
neighbours of a node $v_i$ as $\mathcal{N}(v_i) = \{v_j \,:\, (v_j,v_i)
\in \mathcal{E}\}$, and the set of node features as $\boldsymbol{X} =
\{\boldsymbol{x}_v \in \mathbb{R}^d \,:\, v \in \mathcal{V}\}$ for
some $d \in \mathbb{N}$. On graphs defined as such, the task of
\emph{node classification} consists of learning a function $f:
\mathcal{V} \to \mathcal{Y}$ that maps nodes to labels $y_v \in
\mathcal{Y}$. Typically, node classification is a \emph{transductive}
task, meaning that the learning algorithm has access to the entire
structure of $G$ but only to a subset $\mathcal{V}_\text{labelled}
\subset \mathcal{V}$ of the node labels, and the goal is to predict
the labels on the unlabelled subset $\mathcal{V}_\text{unlabelled} =
\mathcal{V} \setminus \mathcal{V}_\text{labelled}$.

% ----------------------------------------------------------

\subsubsection{Graph neural networks}
\noindent Learning from graph data poses unique challenges such as
handling variable-sized topologies, capturing potentially non-linear
connectivity patterns, and managing cycles. Graph neural networks
(GNNs) \citep{bacciu2023gentle,wu2021comprehensive} facilitate the
adaptive processing of graphs by iteratively refining node
representations (\emph{embeddings}) through neighbourhood aggregations,
thereby progressively increasing the receptive field of the
nodes~\citep{Micheli-NN4G}. In practice, a GNN computes the embedding
of a node $v$ by taking as input its features $\boldsymbol{x}_v$ and
those of its neighbours $\{\boldsymbol{x}_u \,:\, u \in
\mathcal{N}(v)\}$, returning as output a node embedding
$\boldsymbol{h}_v^{(L)} \in \mathbb{R}^{d'}$ (for some $d' \in
\mathbb{N}$) obtained after $L \geq 1$ local message-passing
iterations. Once computed, node embeddings can be used to learn
downstream tasks.

Within the GNN family, convolutional approaches compute node
embeddings by stacking multiple message-passing layers, allowing to
learn tasks in an end-to-end fashion. However, training convolutional
GNNs poses several challenges, including a bias towards graphs with
high homophily~\citep{Zhu2020} and the issue of over-smoothing, which
causes the node embeddings to become indistinguishable as the number
of layers increases~\citep{Chen2020aaai}. Addressing heterophily,
i.e., a prevalence of inter-class edges between nodes (the opposite of
homophily), is related to over-smoothing, as successive
message-passing iterations make the embeddings of neighbouring nodes
more similar~\citep{Yan2022}.

In contrast, recursive GNN approaches~\citep{Scarselli2009} frame the
embedding computation as an iterative map akin to the state transition
function of a dynamical system. Instead of having different
message-passing layers each with its own weights, recursive GNNs apply
the same message-passing function within a layer for $L$ iterations.
For graph-level tasks using a pooled representation of the entire
graph, the recursive map is required to possess contractive dynamics,
i.e., a Lipschitz constant smaller than
$1$~\citep{Tortorella2022ijcnn}, while node-level tasks usually take
advantage of the opposite~\citep{Micheli2023neurocomp}. \textit{Graph
Echo State Networks} (\textbf{GESN}) belong to this class of graph
models, adopting additionally the reservoir computing
paradigm~\citep{Lukosevicius2009csr, Nakajima2021book}, meaning that
the weights of the recursive map are specifically initialized to
satisfy constraints on the Lipschitz constant and frozen, while only
the task prediction layer is trained~\citep{Gallicchio2010ijcnn,
Gallicchio2020gesn}.

% -----------------------------------------------------------

\section{Related work}
\label{sec:related-work}
\noindent \citet{Tang:2010uq} introduced two distinct NQ approaches.
The first employs the \emph{weighted vote Relational Neighbour}
(\textbf{wvRN})~\citep{Macskassy:2003sr} algorithm to classify all
nodes in the graph. It works by computing an initial prediction
$\hat{y}_v^{(0)}$ for all nodes $v \in V$ using a base classifier, and
then updating the predictions as:
\begin{equation}
 \hat{y}_v^{(t)} \gets \arg\max_y \sum_{u \in \mathcal{N}(v)} 
 w_{uv}\mathds{1}[\hat{y}_u^{(t-1)}=y] \; \quad \; t \geq 1
\end{equation}
which corresponds to propagating to the current node the most frequent
label among its neighbours, weighting their vote by the strength of the
connection $w_{uv} \in \mathbb{R}$. Once the propagation has
converged, a quantification algorithm is applied. The second approach
by the same authors, called Link-Based Quantification (LBQ), is out of
the scope of this study since it is only suited for graph-level
quantification and is non-aggregative, i.e., it provides class
prevalence estimations without an intermediate classification step.

\citet{Milli:2015mz} proposed two other methods, \emph{Community
Discovery for Quantification} (\textbf{CDQ}) and \emph{Ego Networks
for Quantification} (\textbf{ENQ}). Both methods initially assign a
class to each unlabelled node $v \in \mathcal{V}_\text{unlabelled}$ and
then apply quantification on top. CDQ employs a community discovery
algorithm to group nodes based on the network's topological structure.
An unlabelled node $v \in \mathcal{V}_\text{unlabelled}$ is assigned the
most frequent class within its community. If $v$ belongs to multiple
communities, two strategies are considered: a frequency-based
strategy, which assigns the class label with the highest relative
frequency, and a density-based strategy, which assigns the most
frequent class of the denser community. The ENQ labeling process
utilizes the concept of ego networks. The ego network of radius $r$ of
a node $v$ is the sub-network that includes $v$, referred to as the
ego, and all nodes within the $r$-hop neighbourhood of $v$. The class
of $v$ is then the most frequent class within its ego network. In the
case of isolated nodes, both algorithms assign the class following the
training distribution. One drawback of the methods discussed above is
that they are not designed to leverage node features. Another major
limitation is that they assume homophily within the graph, since in
the intermediate classification step they assign labels to nodes based
on the labels of their neighbours.

% -----------------------------------------------------------

\section{Method}
\label{sec:method}
\begin{figure}[t]
 \centering \includegraphics[height=2.75cm]{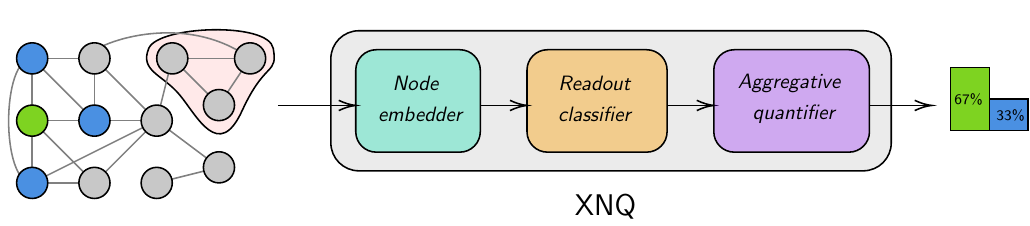}
 \caption{XNQ is composed of three modules applied sequentially.}
 \label{fig:xnq-modules}
\end{figure}

\noindent We start this section by restating the objective of NQ for
clarity. Given a partially labelled graph $G$ with $\mathcal{V} =
\mathcal{V}_\text{labelled} \cup \mathcal{V}_\text{unlabelled}$, our
goal is to produce an estimate $\hat{p}_{\sample}(\oplus)$ of the
proportion of positive nodes in any unlabelled subset $\sample
\subseteq \mathcal{V}_\text{unlabelled}$. To tackle NQ the problem, we
develop the \emph{eXtreme Network Quantifier} (\textbf{XNQ}) model,
referring to its enhanced efficiency and effectiveness. At a high
level, XNQ is composed of three modules applied sequentially:
\begin{enumerate}

\item An unsupervised \emph{node embedder} which computes node
 embeddings leveraging the node features and the graph structure;
 
\item An intermediate \emph{readout classifier} which takes the node
 embeddings as input and computes the node class posterior
 predictions as output;
 
\item A downstream \emph{aggregative quantifier} which aggregates the
 classifier's posterior predictions and estimates the class
 prevalence values.
 
\end{enumerate}
\noindent The three components are shown visually in
Figure~\ref{fig:xnq-modules}. In the following, we describe each
component in detail.

% ----------------------------------------------------------

\subsection{Unsupervised node embedder}
Differently from existing NQ methods, XNQ leverages the node
representations computed by a GNN model in order to exploit
information on input node features as well as the comprehensive graph
topology. To satisfy the requirement of efficiency, XNQ exploits a
reservoir computing GNN such as GESN to embed nodes in an
\textit{unsupervised} and \textit{untrained} fashion, allowing it to
scale to larger networks without requiring a too large fraction of
annotated nodes: as opposed to end-to-end trained GNNs, target nodes
are not used for learning node representations, but only for training
the classifier readout. GESN-based models have proven effective in
solving node classification tasks, reaching state-of-the-art accuracy
on several heterophilic graph benchmarks, while also reducing
computation time compared to fully-trained graph neural
networks~\citep{Micheli2023neurocomp}. Specifically, in XNQ node
embeddings $\boldsymbol{h}_v^{(L)}$ are recursively computed by the
following dynamical system, called the \emph{reservoir}, which
implements an untrained GNN layer, i.e.,
\begin{equation}
 \label{eq:graphesn}
 \boldsymbol{h}_v^{(0)} \gets \boldsymbol{0}, \quad\quad
 \boldsymbol{h}_v^{(\ell)} \gets \tanh\left(\boldsymbol{W}_{\mathrm{in}}\, 
 \boldsymbol{x}_v + \sum_{u \in \mathcal{N}(v)} \boldsymbol{\hat{W}}\, 
 \boldsymbol{h}_{u}^{(\ell-1)} + \boldsymbol{b}_{\mathrm{in}}\right)
\end{equation}
where $\boldsymbol{W}_{\mathrm{in}} \in \mathbb{R}^{d' \times d}$ and
$\boldsymbol{\hat{W}} \in \mathbb{R}^{d' \times d'}$ are the
input-to-reservoir and the reservoir recurrent weights, respectively
($\boldsymbol{b}_{\mathrm{in}} \in \mathbb{R}^{d'}$ is the input
bias). The embedding dimension $d' \in \mathbb{N}$ is a hyperparameter
chosen by model selection, and its value is typically much larger than
the input dimension $d$. Reservoir weights are randomly initialized
from a uniform distribution, and then rescaled to the desired input
range and reservoir spectral radius (also chosen via model selection),
without requiring any training. The number of message-passing
iterations $1 \leq \ell < L$ is set to be larger than the graph
diameter, so as to have a comprehensive receptive field. Of crucial
importance is the Lipschitz constant of the recursive map defined in
Equation~\eqref{eq:graphesn}, which is controlled by setting the
spectral radius of $\boldsymbol{\hat{W}}$, i.e., the largest
eigenvalue modulus $\rho(\boldsymbol{\hat{W}})$. Initializing the
recurrent matrix with $\rho(\boldsymbol{\hat{W}}) < 1 /
\rho(\boldsymbol{A})$, where $\rho(\boldsymbol{A})$ is the graph
spectral radius, implies that the map is contractive and that the
sensitivity of the node embeddings to long-range interactions is
exponentially vanishing~\citep{Micheli2023neurocomp}. Since this
setting leads to over-smoothing, node-level tasks frequently benefit
from recurrent matrix initializations with $\rho(\boldsymbol{\hat{W}})
> 1 / \rho(\boldsymbol{A})$. This holds particularly true for
heterophilic graphs, where sensitivity only to the immediate neighbours
may lead to misleading representations. After being iteratively
computed by Equation~\eqref{eq:graphesn} on the whole graph $G$, the
node embeddings $\boldsymbol{h}_v^{(L)}, \forall v \in \mathcal{V}$
are passed to the readout classifier.

% ----------------------------------------------------------

\subsection{Intermediate readout classifier}\label{sec:intreadout}
\noindent XNQ uses a trained logistic regression readout module to
compute the node predictions. The use of logistic regression is
tightly coupled with choosing a GESN-based embedder, since the
high-dimensional expansion performed by the reservoir usually results
in a linear separation of the embeddings. Consequently, a linear model
can be used to learn the classification rule, further contributing to
making the approach extremely efficient. Specifically, our readout
module takes the node embeddings $\boldsymbol{h}_v^{(L)}$ as input,
and computes a raw posterior probability $\bar{y}_v \in [0,1]$ as
\begin{equation}
 \bar{y}_v \gets 
 \mathrm{sigmoid}\left(\boldsymbol{w}_{\mathrm{out}}^\top\, 
 \boldsymbol{h}_v^{(L)} + b_\mathrm{out}\right)
\end{equation}
where $\boldsymbol{w}_{\mathrm{out}} \in \mathbb{R}^{d'}$ are
learnable weights and $b_\mathrm{out} \in \mathbb{R}$ is a learnable
bias. Once the readout has been learned, the output probabilities are
calibrated and passed to the downstream quantifier. Calibration
entails adjusting the output probabilities such that $\Pr(Y = \oplus
\,|\, \bar{Y} = \bar{y}_v) \approx \bar{y}_v$, where $\bar{Y}$ is a
random variable that ranges over $[0,1]$. In other words, by
calibrating the output of the readout we are adjusting the predicted
probabilities to approximately match the observed class frequencies.
In our implementation, calibration is achieved via Platt scaling~\citep{Platt:2000fk}, i.e., by transforming the raw posterior probabilities as
\begin{equation}
 \hat{y}_v \gets \frac{1}{1 + e^{(a\,\bar{y}_v+b)}}
\end{equation}
where $a, b \in \mathbb{R}$ are parameters learned with maximum likelihood. 
In preliminary experiments we also tested, as an alternative calibration method, Bias-Corrected Temperature Scaling (BCTS), which in related work~\citep{Alexandari:2020dn} had shown to improve quantification performance; however, in our experiments, Platt scaling revealed superior, and we thus decided to stick to it. The readout classifier is
trained and calibrated using the labelled node embeddings
$\boldsymbol{h}_v^{(L)},\, \forall v \in \mathcal{V}_\text{labelled}$.
Then, we use it to predict the unlabelled nodes in
$\mathcal{V}_\text{unlabelled}$, obtaining a set of calibrated
posterior probabilities 
$\hat{\Ptr}(\oplus\,|\, \boldsymbol{h}_v^{(L)})$, for all $v \in \mathcal{V}_\text{unlabelled}$.

Note that the calibration property is defined upon a set (or a distribution), meaning that a classifier that has been calibrated for a training distribution will hardly generate well-calibrated posteriors for a different test distribution, especially when the training and test distributions are related by PPS, since its underlying assumptions imply $\Ptr(\oplus\,|\, \boldsymbol{h}_v^{(L)})\neq\Pte(\oplus\,|\, \boldsymbol{h}_v^{(L)})$. How to properly adapt the (estimated) posteriors to the (unknown) test prior is the subject of the next section.

% ----------------------------------------------------------

\subsection{Downstream aggregative quantifier}
\label{sec:sld}
The goal of XNQ's downstream quantifier is to output the
desired estimate $\hat{p}_\sample(\oplus)$ in an unlabelled subset
$\sample \subseteq \mathcal{V}_\text{unlabelled}$ using the
observed training proportion $p_\text{labelled}(\oplus)$ and the
estimated posterior probabilities $\hat\Ptr(\oplus\,|\,\boldsymbol{h}_v^{(L)}),\,
\forall v \in \sample$ as inputs. We do so by adapting the
\textit{Saerens-Latinne-Decaestecker} method~\citep{Saerens:2002uq}
(\textbf{SLD}) to our setting. The rationale behind the choice are its
strong performance in the challenge of quantification on data affected
by PPS and its desirable theoretical guarantees. Indeed, SLD is proven
to be \emph{Fisher-consistent} under PPS~\citep{Tasche:2017ij}, i.e.,
its class prevalence estimates are guaranteed correct under PPS if
computed on the whole populations of interest (instead of the limited
samples $\mathcal{V}_\text{unlabelled}$ and $\sample$). Basically,
SLD is an instance of the Expectation-Maximization (EM) algorithm.
Initially, the prevalence estimates are set to
$\hat{p}_\sample^{(0)}(\oplus) \gets p_\text{labelled}(\oplus)$.
Then, two mutually recursive steps are iterated (for $k \geq 1$):
\begin{itemize}

\item the \textit{E-step}: The posterior probability $\Pte(\oplus\,|\,\boldsymbol{h}_v^{(L)})$ is approximated by scaling the estimated posterior
 $\hat\Ptr(\oplus\,|\,\boldsymbol{h}_v^{(L)})$ by the ratio
 between the previous estimate $\hat{p}_\sample^{(k-1)}(\oplus)$
 and the initial estimate $\hat{p}_\sample^{(0)}(\oplus)$, and
 re-normalized. This tunes the posterior probabilities towards the
 current class prevalence estimate.
 
\item the \textit{M-step}: The current estimate
 $\hat{p}_\sample^{(k)}(\oplus)$ is updated as the average of the current posterior estimates obtained in the \textit{E-step}. This tunes the class prevalence
 estimate towards the rescaled posterior probabilities.
 
\end{itemize}
\noindent The process is repeated until the estimate remains stable
through successive iterations, at which point the final estimate
$\hat{p}_\sample(\oplus)$ is returned.

% -----------------------------------------------------------

\section{Experiments and discussion}
\label{sec:experiments}
\noindent We compare our proposed XNQ against the previous literature
methods described in Section~\ref{sec:related-work} (wvRN, CDQ, ENQ)
with the exception of LBQ as it cannot operate quantification at the
node level. For each baseline, we optimize the downstream quantification method using the same experimental protocol as XNQ.

% -----------------------------------------------------------

\subsection{Experimental protocol}
\label{sec:experimental-protocol}

% -----------------------------------------------------------

\subsubsection{Datasets}
\label{sec:Datasets}
\noindent We select five publicly available datasets from the node
classification literature, adapting them to NQ:
\begin{itemize}

\item \textit{Cora}: a citation network first introduced
 in~\citep{McCallum2000cora}. Nodes in the graph are scientific
 publications, links are citations, and the task consists of picking the
 scientific field the publication belongs to from a set of 7 such fields~\citep{yang2016cora}. Each
 node has textual features extracted from the abstract.
 
\item \textit{Genius}: the social network from a crowd-sourced song
 annotation platform~\citep{Lim2021}. Links represent friendship
 between users, whose accounts are classified as regular or spam.
 
\item \textit{Questions}: a network from the question-answering
 website Yandex Q where nodes are users, and links specify whether
 one user answered the other's questions within a certain time
 span~\citep{platonov2023questions-tolokers}. User accounts are
 classified into active or inactive.
 
\item \textit{Tolokers}: a network from the Toloka
 platform~\citep{Likhobaba2023toloker}, where nodes are workers who
 participated in crowd-sourcing projects, and links specify mutual
 participation. The positive class corresponds to banned
 users~\citep{platonov2023questions-tolokers}.
 
\item \textit{Twitch-DE}: a social network of German users from the
 Twitch streaming platform~\citep{Rozemberczki2021twitch}. The
 positive class corresponds to adult content profiles.
 
\end{itemize}
\noindent All datasets except Cora are real-world networks that range
in size up to 400K nodes and 1M edges, and present binary
classification tasks characterized by a high degree of heterophily
(i.e., a large fraction of inter-class edges). Cora is instead a multi-class dataset (with 7 classes); from it we have derived a binary 1-vs-rest
classification task by picking class 2 as the ``positive class'' $\oplus$, to serve as a high-homophily control case. Table~\ref{tab:datasets} reports the main statistics of these datasets, including the prevalence of the positive class and the
class-adjusted homophily~\citep{Lim2021}, which measures the prevalence of intra-class edges in the presence of class imbalance (higher values indicate higher homophily).

\begin{table}[t]
 \caption{Main characteristics of the datasets considered in this
 study.}
 \label{tab:datasets}
 \centering \scriptsize
 \begin{tabular}{lrrrrr}
 \toprule
 & \multicolumn{1}{c}{\textbf{Cora-2}} & 
 \multicolumn{1}{c}{\textbf{Genius}} 
 & \multicolumn{1}{c}{\textbf{Questions}} & 
 \multicolumn{1}{c}{\textbf{Tolokers}} 
 & \multicolumn{1}{c}{\textbf{Twitch-DE}} \\
 \midrule
 Number of nodes & 2,708 & 421,961 & 48,921 & 11,758 & 9,498 \\
 Number of edges & 5,429 & 984,979 & 307,080 & 1,038,000 & 315,774 \\
 Node input features & 1,433 & 12 & 301 & 10 & 128 \\
 Prevalence of the $\oplus$ class & 0.15 & 0.20 & 0.03 & 0.02 & 0.61 \\
 Class-adjusted homophily & 0.89 & 0.22 & 0.18 & 0.08 & 0.17 \\
 \bottomrule
 \end{tabular}
\end{table}

% -----------------------------------------------------------

\subsubsection{Evaluation setting}
\noindent We use 5-fold cross-validation (CV) to estimate the
out-of-sample AE. The set of nodes is divided in 5 equally-sized,
disjoint partitions. In turn, 4 folds are used as development set,
while the remaining fold is reserved for test evaluation. In each
development set, $25\%$ of the data is held out as validation set to
perform model selection via grid search; an additional $12.5\%$ of the
data is held out for calibration (which is required by quantification
methods such as PCC, PACC, HDy, DyS, SLD); the remaining $62.5\%$ of
the development data is used for training. The model configuration
achieving the lowest AE estimated from $100$ instances of the
artificial prevalence protocol (Section~\ref{sec:quantification})
applied on the validation set is then evaluated on the corresponding
test set. The average AE score estimated via the APP protocol on each
of the 5 test folds is reported, along with the standard deviation.
We remark that all combinations of models, hyper-parameters and
quantification methods are trained and evaluated on the same data
splits to ensure fairness. Further information allowing to reproduce the experiments is provided in the code repository.

% -----------------------------------------------------------

\subsubsection{Hyperparameters}\label{sec:hyperparameters}
XNQ hyperparameters were optimized using randomized
search. Specifically, we sampled and evaluated 100 configurations
choosing among the following: embedding size (sampled uniformly from the set $\{512, 1024, 2048, 4096\}$ (except for Genius, in which we restricted the exploration to $\{512,1024\}$ due to hardware limitations), recurrent scale (sampled
log-uniformly from the interval $[1, 25]$), input scale (sampled
log-uniformly from the interval $[0.1, 1]$), and readout regularization
coefficient (sampled log-uniformly from the interval
$[10^{-2},10^3]$).

The existing methods with which we compare to were tuned with
grid search as follows. For CDQ, we tuned the merge strategy, choosing
between frequency-based and density-based. These are the only
hyperparameters of this method. For ENQ, we tuned the hops to extract
the ego-networks from the nodes, choosing from the set $\{1,2\}$, limiting the exploration to either 1 or 2 hops (higher values revealed computationally intractable). Since both methods only assign crisp (non-probabilistic) labels, they choose the quantifier from the set $\{\text{CC, ACC}\}$. For wvRN, we tuned the
number of iterations choosing from the set $\{1,2,3,4,5\}$, while the
quantifier was chosen from the set $\{\text{CC, ACC, PCC, PACC, HDy, DyS, SLD}\}$.

The ablation baselines were tuned via grid search as
follows. For LR, we optimized the regularization coefficient in a grid
of 10 log-uniformly distributed values in the interval
$[10^{-2},10^3]$. For the convolutional GNNs, we optimized the
following hyperparameters: number of layers (from the set $\{2,3,4\}$)
and embedding dimension (from the set $\{128, 256\}$). Both LR and the
convolutional GNNs also optimized the quantifier choice by inspecting the
set $\{\text{CC, ACC, PCC, PACC, HDy, DyS, SLD}\}$.

In order to guarantee a fair comparison in terms of resource constraints, we granted a budget of 24 hours of computation to every model for exploring the hyperparameter space. Therefore, we are confident that our model selection setup is sufficiently
fair given the computational constraints.

% -----------------------------------------------------------

\subsection{Results}
\label{sec:results}
\noindent Table~\ref{tab:results} reports the average AE across the 5
CV folds using the APP testing protocol. XNQ achieves the lowest AE
average across all 5 datasets, outperforming all existing NQ methods
to date. The most significant improvement with respect to previous
state-of-the-art (a $90.48\%$ reduction in AE) is observed on the
Genius dataset, which has the largest number of nodes. On the Tolokers
dataset, characterized by the highest number of edges and the least
homophily, the improvement is $59.49\%$. For the Questions and
Twitch-DE datasets, with a positive class prevalence of $0.03$ and
$0.61$ respectively (the former highly unbalanced, the latter mostly
balanced), the relative improvements are $84.11\%$ and $33.33\%$.
Even on the Cora-2 dataset, whose extremely high homophily should be
congenial for the baselines, XNQ demonstrates a $48.27\%$ error
reduction with respect to the runner-up method. These results
underscore XNQ's robust performance across graphs of varying sizes,
training prevalence values, and heterophily levels.
The Bayesian analysis reported in Appendix~\ref{app:statistical-significance} confirms XNQ as the significantly best method for NQ.

For completeness, in Table~\ref{tab:results-rae} we also report the results of the evaluation in terms of \textit{relative absolute error} (RAE). In general, the two measures may paint two different pictures, since it is often the case that method A is better than method B at AE while the opposite is true at RAE \citep{Esuli:2022wf}.
orEven using a different evaluation measure, XNQ manages to outperform all the other methods in almost all cases, the only exception being the Twitch-DE dataset, where it achieves the second-best result, with ENQ the best performer. The statistical significance analysis of Appendix~\ref{app:statistical-significance} confirms this picture.

\begin{figure}[!ht]
 \centering
 \begin{subfigure}{.5\textwidth}
 \centering \includegraphics[width=\textwidth]{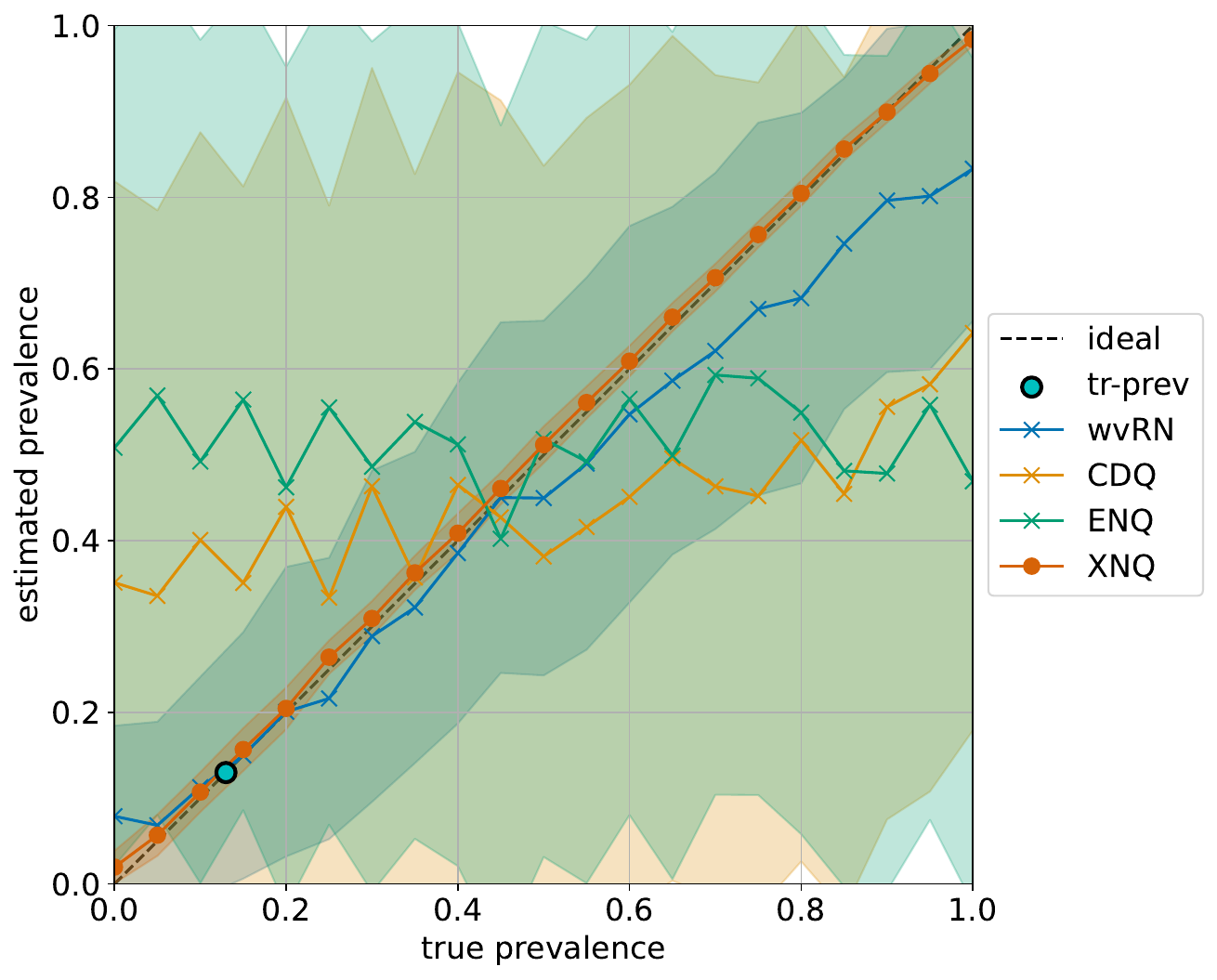}
 \subcaption{Genius}
 \label{fig:diagonal-genius}
 \end{subfigure}%
 \begin{subfigure}{.5\textwidth}
 \centering \includegraphics[width=\textwidth]{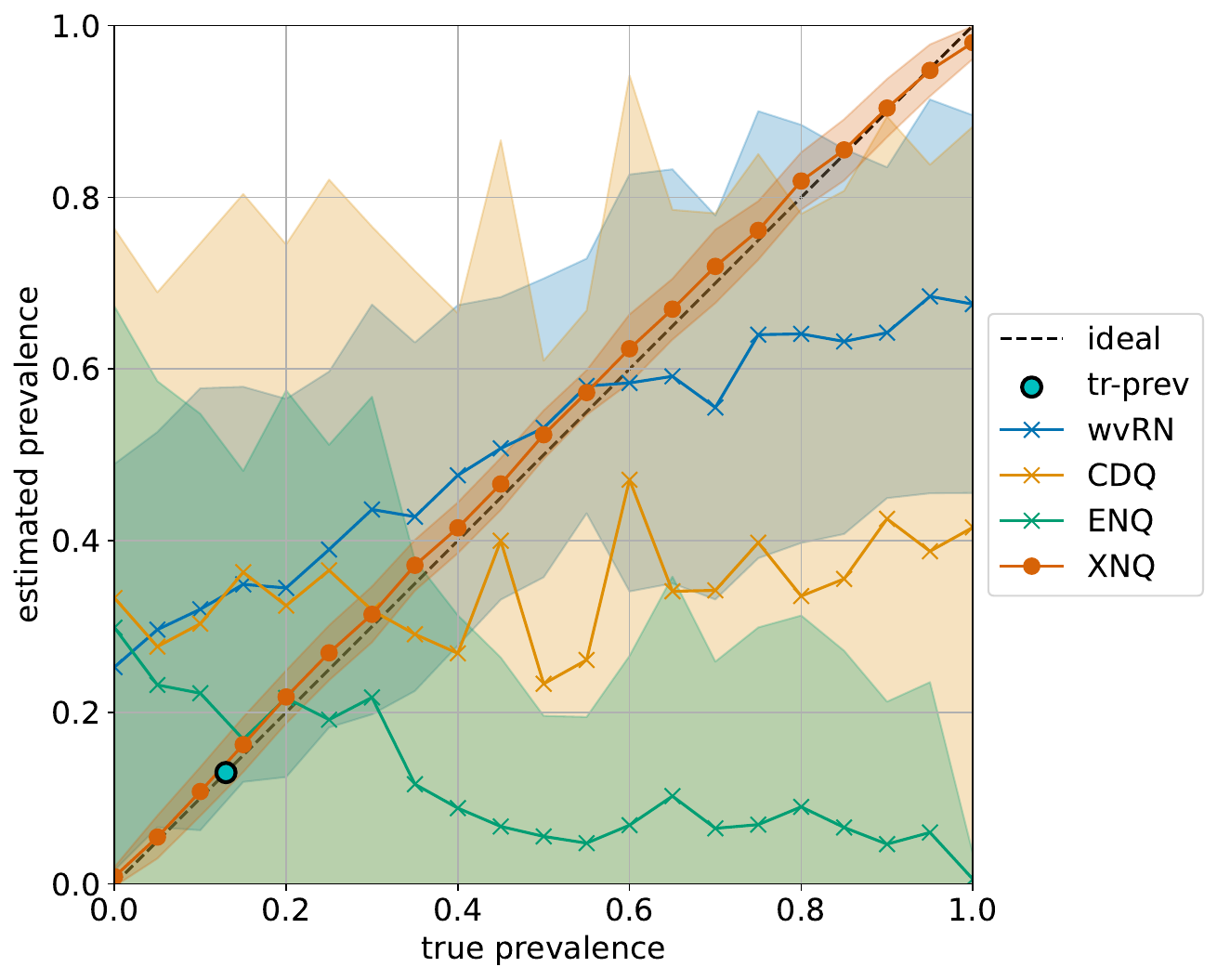}
 \subcaption{Questions}
 \label{fig:diagonal-questions}
 \end{subfigure}
 \begin{subfigure}{.5\textwidth}
 \centering \includegraphics[width=\textwidth]{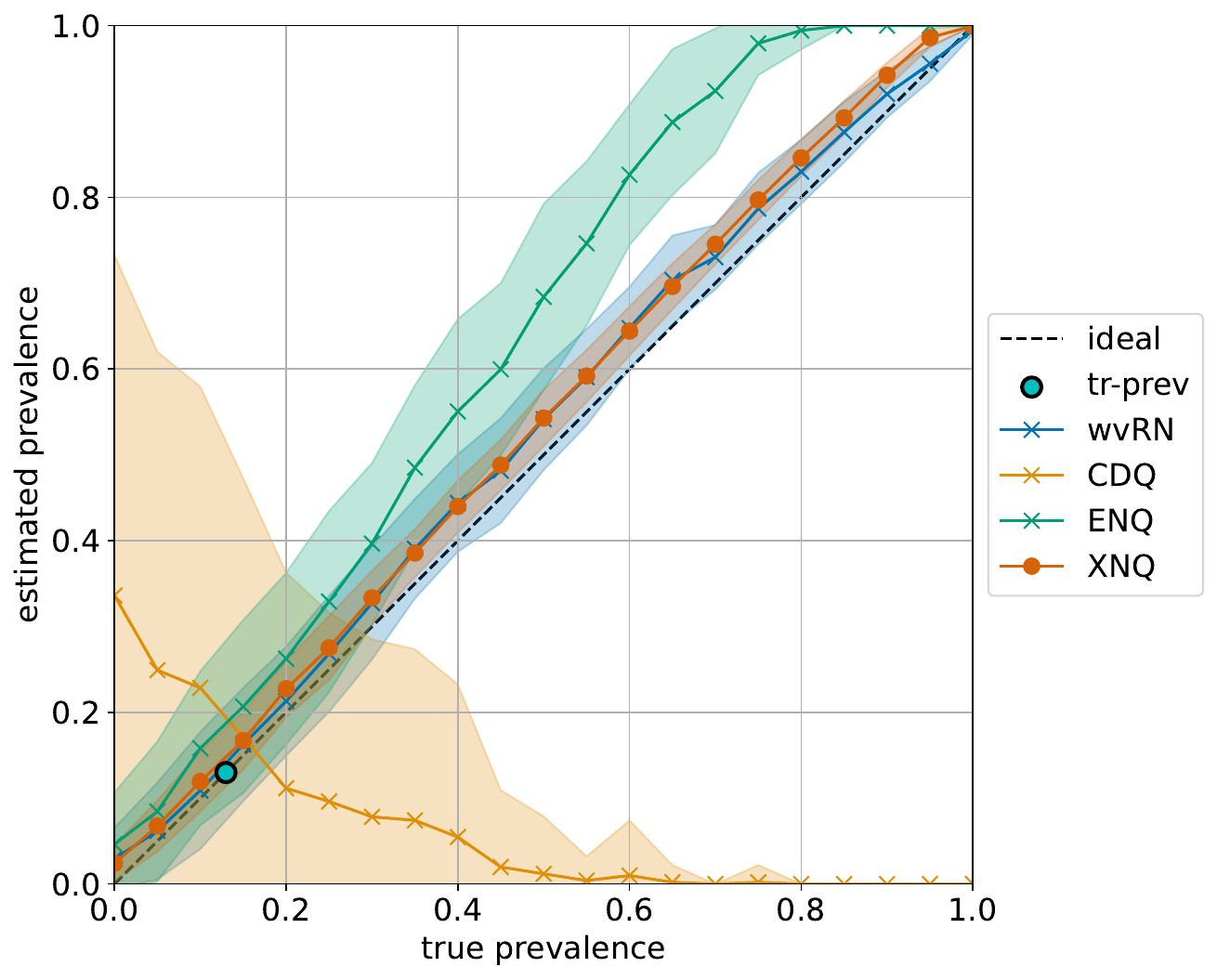}
 \subcaption{Tolokers}
 \label{fig:diagonal-tolokers}
 \end{subfigure}%
 \begin{subfigure}{.5\textwidth}
 \centering \includegraphics[width=\textwidth]{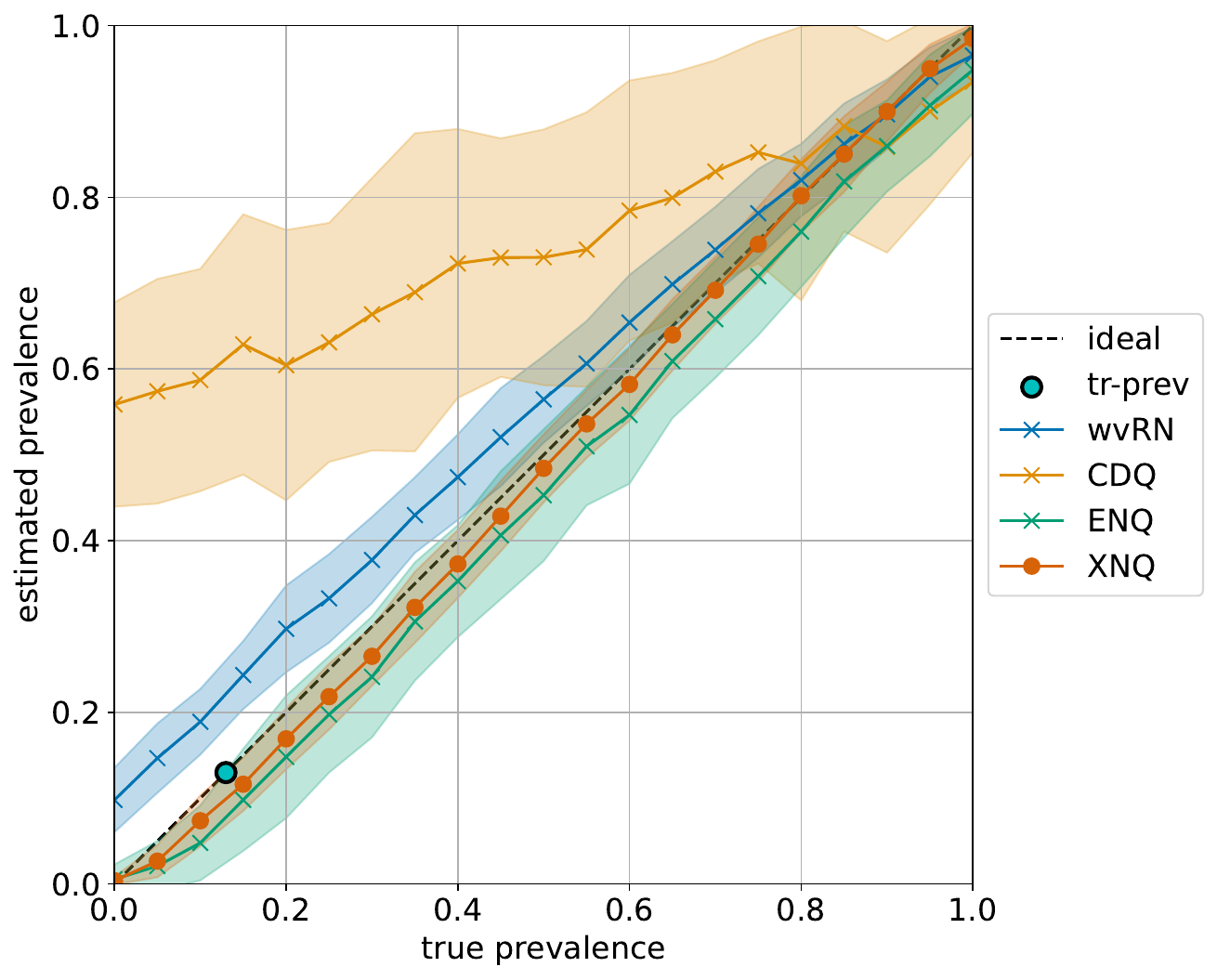}
 \subcaption{Twitch-DE}
 \label{fig:diagonal-twitch}
 \end{subfigure}
 \caption{Diagonal plots compare XNQ against other NQ
 baselines for different test class prevalence values generated via the APP. Shaded
 bands represent standard deviations.}
 \label{fig:diagonal-plot}
\end{figure}

\begin{table}[t]
 \caption{Results of the evaluation (best performance in
 \textbf{boldface}, second-best performance
 \underline{underlined}). Reported results are the 5-fold CV AE
 averages on the test folds. Row ``Improvement'' reports the
 improvement (error reduction) of XNQ with respect to the second-best
 performance.}\label{tab:results}
 \centering

 \begin{tabular}{lcccccc} \toprule \textbf{Method} & \textbf{Cora-2} &
 \textbf{Genius} & \textbf{Questions} & \textbf{Tolokers} &
 \textbf{Twitch-DE}\\ \midrule wvRN & {$0.037 \pm 0.020$} &
 {$\underline{0.147} \pm 0.005$} & {$0.214 \pm 0.040$} &
 {$\underline{0.079} \pm 0.036$}
 & {$\underline{0.060} \pm 0.011$} \\
 CDQ & {$0.100 \pm 0.045$} & {$0.409 \pm 0.017$} & {$0.354 \pm 0.098$} &
 {$0.358
 \pm
 0.141$}
 & {$0.270 \pm 0.067$} \\
 ENQ & {$\underline{0.029} \pm 0.008$} & {$0.476 \pm 0.001$} &
 {$\underline{0.159}
 \pm
 0.007$} &
 {$0.099
 \pm
 0.036$}
 & {$0.089 \pm 0.036$} \\
 \midrule XNQ & {$\mathbf{0.015} \pm 0.011$} & {$\mathbf{0.015} \pm
 0.001$} &
 {$\mathbf{0.034}
 \pm 0.007$} &
 {$\mathbf{0.032}
 \pm
 0.010$}
 & {$\mathbf{0.040} \pm 0.010$} \\
 Improvement & {$-48.27\%$} & {$-90.48\%$} & {$-84.11\%$} &
 {$-59.49\%$}
 & {$-33.33\%$} \\
 \bottomrule
 \end{tabular}
\end{table}
Figure~\ref{fig:diagonal-plot} shows the diagonal plots on the four
heterophilic datasets (Cora-2 is omitted as all methods have similarly
looking diagonal plots). It can be noticed that while all baselines
have an erratic behaviour depending on the dataset, XNQ consistently
constantly maintains strong performance, regardless of the
characteristics of the dataset to which it is being applied,
approaching the ideal performance line that bisects the plotting
space.

\begin{table}[t]
 \caption{Results of the evaluation (best performance in
 \textbf{boldface}, second-best performance
 \underline{underlined}). Reported results are the 5-fold CV RAE
 averages on the test folds. The row ``Improvement'' reports the
 improvement (error reduction) of XNQ with respect to the second-best
 performance.}\label{tab:results-rae}
 \centering

 \begin{tabular}{lrrrrr} \toprule \textbf{Method} & \multicolumn{1}{c}{\textbf{Cora-2}} &
 \multicolumn{1}{c}{\textbf{Genius}} & \multicolumn{1}{c}{\textbf{Questions}} & \multicolumn{1}{c}{\textbf{Tolokers}} &
 \multicolumn{1}{c}{\textbf{Twitch-DE}}\\ 
 \midrule 
 wvRN & $0.976 \pm 1.758$ & $ \underline{2.605} \pm 0.539$ & $12.718 \pm 7.930$ & $ 6.151 \pm 8.407$ & $3.465 \pm 3.070$ \\
 CDQ & $1.679 \pm 1.101$ & $18.304 \pm 2.047$ & $25.152 \pm 2.702$ & $20.150 \pm 12.42$ & $11.234 \pm 4.352$ \\
 ENQ & $\underline{0.751} \pm 0.987$ & $13.246 \pm 7.278$ & $ \underline{4.775} \pm 2.473$ & $ \underline{2.903} \pm 1.398$ & $\mathbf{2.316} \pm 0.949$ \\
 \midrule 
 XNQ & $\mathbf{0.241} \pm 0.374$ & $\mathbf{0.417} \pm 0.025$ & $\mathbf{2.161} \pm 1.020$ & $\mathbf{0.441} \pm 0.253$ & $\underline{3.254} \pm 1.508$ \\
Improvement & \multicolumn{1}{c}{$-67.95\%$} & \multicolumn{1}{c}{$-83.99\%$} & \multicolumn{1}{c}{$-54.74\%$} & \multicolumn{1}{c}{$-84.80\%$} & \multicolumn{1}{c}{--} \\
 \bottomrule
 \end{tabular}
\end{table}
%

% -----------------------------------------------------------

\subsection{Ablation study}
\label{sec:ablations}
\noindent To gain deeper insights into XNQ's performance and to
validate our design, we conduct ablation experiments by modifying the
underlying components of XNQ, following the exact setup described in
Section~\ref{sec:experimental-protocol}. Specifically, we study how
performance is affected by varying the component of XNQ that computes
the node embeddings, and the component corresponding to the downstream
quantification method.

% -----------------------------------------------------------

\subsubsection{Node embedder ablation}
\noindent We replace the original node embedder of XNQ with multiple
layers of Graph Convolutional Network (GCN)~\citep{kipf2017gcn}, Graph
Attention Network (GAT)~\citep{velickovic2018gat}, and Graph
Isomorphism Network (GIN)~\citep{xu2018gin}, with the number of layers
treated as a hyper-parameter. These methods are briefly recapped in
Appendix~\ref{app:conv-gnn}. Following best practices to evaluate
GNNs~\citep{errica2020fair}, we also include a network-agnostic
variant model which applies a Logistic Regression (LR) readout
directly to the node features without considering the graph structure.
The results, shown in the upper block of Table~\ref{tab:ablation},
clearly indicate that XNQ exploits the graph structure better than the
baselines, with superior performances. Surprisingly, the LR baseline
performs better than the convolutional GNNs, which we attribute to the
known issue of these models being difficult to calibrate
properly~\citep{teixeira2019miscalibration}, making them less suitable
for quantification tasks.

\begin{table}[t!]
 \caption{Results of ablating the two main components of XNQ (best
 performance in \textbf{boldface}, second-best performance
 \underline{underlined}). Reported results are the 5-fold CV AE
 averages ($\pm$ standard deviation) obtained on the test
 folds.} \label{tab:ablation} \centering

 \begin{tabular}{llcccccc} \toprule \textbf{Ablation} &
 \textbf{Method} & \textbf{Cora-2} & \textbf{Genius} &
 \textbf{Questions} & \textbf{Tolokers} & \textbf{Twitch-DE}\\
 \midrule
 
 Node & LR
 & {$\underline{.032} \pm .023$} 
 & {$\underline{.018} \pm .001$} 
 & {$\underline{.088} \pm .013$} 
 & {$\underline{.044} \pm .011$} 
 & {$.057 \pm .010$} \\
 
 Embedder & GCN
 & {$.071 \pm .034$} 
 & {$.050 \pm .008$} 
 & {$.106 \pm .035$} 
 & {$.057 \pm .027$} 
 & {$\underline{.045} \pm .012$} \\
 
 & GAT
 & {$.059 \pm .023$} 
 & {$.032 \pm .004$} 
 & {$.096 \pm .041$} 
 & {$.099 \pm .111$} 
 & {$.055 \pm .016$} \\
 
 & GIN
 & {$.075 \pm .014$} 
 & {$.111 \pm .124$} 
 & {$.107 \pm .032$} 
 & {$.051 \pm .020$} 
 & {$.064 \pm .030$} \\
 
 \midrule
 
 Quantifier & CC
 & {$.048 \pm .011$} 
 & {$.086 \pm .002$} 
 & {$.429 \pm .010$} 
 & {$.265 \pm .008$} 
 & {$.203 \pm .007$} \\
 
 & ACC
 & {$\underline{.022} \pm .012$} 
 & {$.017 \pm .000$} 
 & {$.108 \pm .018$} 
 & {$.045 \pm .016$} 
 & {$.056 \pm .013$} \\
 
 & PCC
 & {$.043 \pm .012$} 
 & {$.217 \pm .000$} 
 & {$.410 \pm .009$} 
 & {$.235 \pm .002$} 
 & {$.221 \pm .004$} \\
 
 & PACC
 & {$.026 \pm .012$} 
 & {$\mathbf{.014} \pm .000$} 
 & {$\underline{.068} \pm .021$} 
 & {$\underline{.041} \pm .010$} 
 & {$.048 \pm .011$} \\

 & HDy
 & {$.037 \pm .019$} 
 & {$\mathbf{.014} \pm .000$} 
 & {$.076 \pm .043$} 
 & {$.054 \pm .030$} 
 & {$.070 \pm .012$} \\

 & DyS
 & {$.023 \pm .014$} 
 & {$\mathbf{.014} \pm .000$} 
 & {$.081 \pm .039$} 
 & {$\mathbf{.032} \pm .005$} 
 & {$\underline{.044} \pm .010$} \\
 
 \midrule
 
 Reference & XNQ
 & {$\mathbf{.015} \pm .011$} 
 & {$\underline{.015} \pm .001$} 
 & {$\mathbf{.034} \pm .007$} 
 & {$\mathbf{.032} \pm .010$} 
 & {$\mathbf{.040} \pm .010$} \\
 
 \bottomrule
 \end{tabular}
 \footnotetext{Note: The XNQ results are taken from Table
 \ref{tab:results}.}
\end{table}

% -----------------------------------------------------------

\subsubsection{Quantifier component ablation}
\noindent We replace SLD with different aggregative quantification
methods described in detail in
Appendix~\ref{app:quantification-methods}. According to the results in the
bottom block of Table~\ref{tab:ablation}, XNQ achieves the best
performance in 4 out of 5 cases and secures the second-best
performance in the Genius dataset, with only a tiny margin separating
it from the top methods (PACC, HDy, and DyS). This result confirms
that CC is a completely inadequate method for NQ, which requires
powerful downstream quantifiers. All the other results align with the
existing literature on quantification, where SLD-based algorithms
consistently rank among the top performers, thereby justifying our
decision to integrate an expectation-maximization quantifier into XNQ.

% -----------------------------------------------------------

\subsection{Efficiency considerations}
\label{sec:efficiency}

% -----------------------------------------------------------

\subsubsection{Computation time}
\begin{figure}[ht!]
 \centering \includegraphics[width=\textwidth]{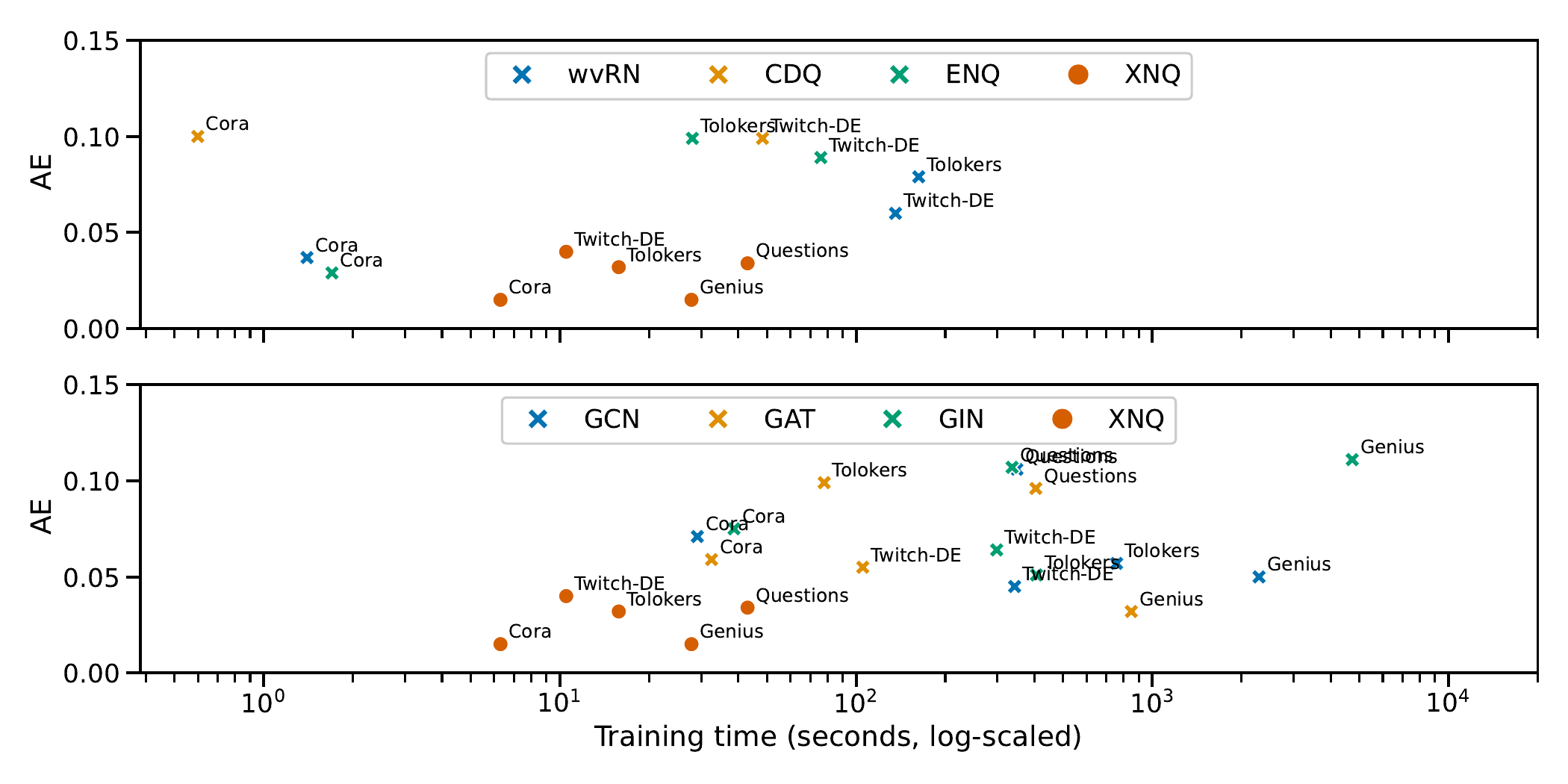}
 \caption{The trade-off between error (AE, $y$ axis) and average time
 (in seconds, log-scaled $x$ axis) required to train the most expensive
 hyperparameter configuration of the models. Only results with AE
 $\leq 0.125$ are shown to avoid cluttering. XNQ occupies the ``sweet
 spot'' close to the origin, where methods are both efficient and
 effective.}
 \label{fig:timing}
\end{figure}

\noindent XNQ turns out to be computationally more efficient than the
other graph-learning methods. To support this claim, for each dataset
we plot on the log-scaled $y$ axis of Figure~\ref{fig:timing} the
average time (over 5 folds) required by each model to train the most
expensive hyper-parameter configuration on a single NVidia V100
GPU. In all cases, XNQ is faster to train than the alternatives, up to
more than one order of magnitude faster for the Genius dataset, taking
less than a single minute to process the network's 400K nodes. Notice
that while the baselines are faster than XNQ in the Cora-2 dataset (top
figure), their AE is worse than the one obtained by XNQ.

% -----------------------------------------------------------

\subsubsection{Data annotation}
\begin{figure}[ht!]
 \centering \includegraphics[width=5cm]{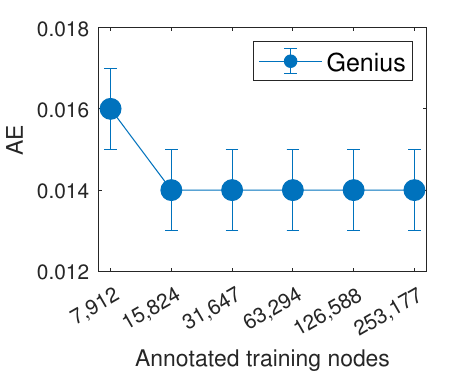}
 \caption{The quantification error of XNQ remains consistently low on
 Genius as the fraction of annotated data is reduced up to
 $\approx2\%$.}
 \label{fig:data-annotation}
\end{figure}

\noindent A particularly onerous challenge in dealing with large-scale
networks is providing enough annotated data to feed to the learning
methods, as it often requires human intervention in real-world
scenarios. Such an example may be users in a social network answering
a preference survey, which is then used as the annotated data to
quantify user preferences within particular communities. In
Figure~\ref{fig:data-annotation}, we show that the AE of XNQ on Genius
(the largest network) stays low for increasingly smaller fractions of
annotated training data (down to less than $2\%$ of nodes), showcasing
its scalability to scarcer annotated data scenarios.

% -----------------------------------------------------------

\subsection{Extension to the multi-class setting}\label{sec:multi-class}
\noindent To demonstrate the versatility of XNQ beyond the scope
of binary quantification, we here present the results of an
evaluation in a multi-class quantification setting. From an
operational point of view, adapting from the binary to the multi-class
scenario amounts to replacing the readout module with a multinomial LR
classifier, and replacing the downstream quantifiers with their
multi-class counterparts. Crucially, most quantification methods we have discussed are natively multi-class,\footnote{The only exceptions are DyS and HDy, although multi-class extensions have been proposed in the literature.} making the
extension trivial. We thus conducted an evaluation on the
following datasets:
\begin{itemize}

\item \textit{Cora-Multi}: this is the same Cora dataset we described
 in Section~\ref{sec:Datasets}, but this time we handle all 7 classes
 simultaneously. The class-adjusted homophily is 0.77.
 
\item \textit{Amazon}: a product co-purchasing network where nodes are
 products, and edges connect co-occurring purchases
 \citep{platonov2023questions-tolokers}. The task is to predict the
 average rating (4 classes) given to a product by reviewers. The
 graph has 29,492 nodes (with 300 features each) and 182,100 edges, and
 a class-adjusted homophily of 0.13.
 
\item \textit{Flickr}: a graph where nodes are images uploaded to the
 Flickr online community. Edges connect nodes where respective images
 share some common properties such as same geographic location
 \citep{McAuley2012}. Node features are 500-dimensional bag-of-word
 vectors extracted from image annotations. The task is to classify
 each image into a semantic category (one of 7 classes). The dataset
 has 89,250 nodes and 899,756 edges, with a class-adjusted homophily of
 0.07.
 
\end{itemize}
\noindent We compare XNQ to the same baselines as in
Table~\ref{tab:results}, using the same experimental setup described
in Section~\ref{sec:experiments}. The results are presented in
Table~\ref{tab:results-multi} (for the AE metric) and Table~\ref{tab:results-multi-rae} (for the RAE metric): similarly to what observed in the
binary quantification setup, XNQ consistently outperforms all the
existing baselines in the three datasets. These results underscore the
effectiveness of XNQ and broaden its applicability to NQ tasks having
node label cardinality $>2$. A Bayesian analysis to assess the significance of the results is presented in Appendix~\ref{app:statistical-significance}, confirming the above trend.

\begin{table}[t]
 \caption{Results of the evaluation on multi-class datasets
 (best performance in \textbf{boldface}, second-best performance
 \underline{underlined}). Reported results are the 5-fold CV AE
 averages on the test folds. The row ``Improvement'' reports the improvement (error reduction) of XNQ with respect to the second-best
 performance.}\label{tab:results-multi}
 \centering
 \begin{tabular}{lccc} 
 \toprule \textbf{Method} & \textbf{Cora-Multi} & \textbf{Amazon} & \textbf{Flickr}\\ 
 \midrule 
 wvRN & {$\underline{0.023} \pm 0.006$} & {$\underline{0.108} \pm 0.017$} & {$0.201 \pm 0.066$} \\
 CDQ & {$0.076 \pm 0.015$} & {$0.111 \pm 0.003$} & {$0.145 \pm 0.001$} \\
 ENQ & {$0.028 \pm 0.004$} & {$0.116 \pm 0.016$} & {$\underline{0.106} \pm 0.006$} \\
 \midrule 
 XNQ & {$\mathbf{0.012} \pm 0.002$} & {$\mathbf{0.065} \pm 0.007$} & {$\mathbf{0.075} \pm 0.006$} \\
Improvement & {$-47.82\%$} & {$-39.81\%$} & {$-29.24\%$} \\
 \bottomrule
 \end{tabular}
\end{table}

\begin{table}[t]
 \caption{Results of the evaluation on multi-class datasets
 (best performance in \textbf{boldface}, second-best performance
 \underline{underlined}). Reported results are the 5-fold CV RAE
 averages on the test folds. The row ``Improvement'' reports the improvement (error reduction) of XNQ with respect to the second-best
 performance.}\label{tab:results-multi-rae}
 \centering
 \begin{tabular}{lccc} 
 \toprule 
 \textbf{Method} & \textbf{Cora-Multi} & \textbf{Amazon} & \textbf{Flickr}\\ 
 \midrule 
 wvRN & {$\underline{0.417} \pm 0.095$} & {$\underline{1.424} \pm 0.104$} & {$\underline{1.710} \pm 0.112$} \\
 CDQ & {$1.189 \pm 0.210$} & {$1.576 \pm 0.360$} & {$2.986 \pm 0.521$} \\
 ENQ & {$0.459 \pm 0.067$} & {$1.521 \pm 0.188$} & {$1.884 \pm 0.193$} \\
 \midrule 
 XNQ & {$\mathbf{0.312} \pm 0.025$} & {$\mathbf{1.330} \pm 0.273$} & {$\mathbf{1.588} \pm 0.082$} \\
Improvement & {$-25.17\%$} & {$-6.60\%$} & {$-7.13\%$} \\
 \bottomrule
 \end{tabular}
\end{table}

% -----------------------------------------------------------

\subsection{Application scenarios}\label{sec:applications}

In many real-world scenarios, users are not interested in classifying individual items, but in estimating the prevalence of some characteristics in a subset of the general population.
In this section, we present some application scenarios where XNQ can be employed to address some real-world tasks.

% -----------------------------------------------------------

\paragraph{Quantifying bots and malicious users in social networks}

\begin{figure}[ht!]
 \centering \includegraphics[width=\textwidth]{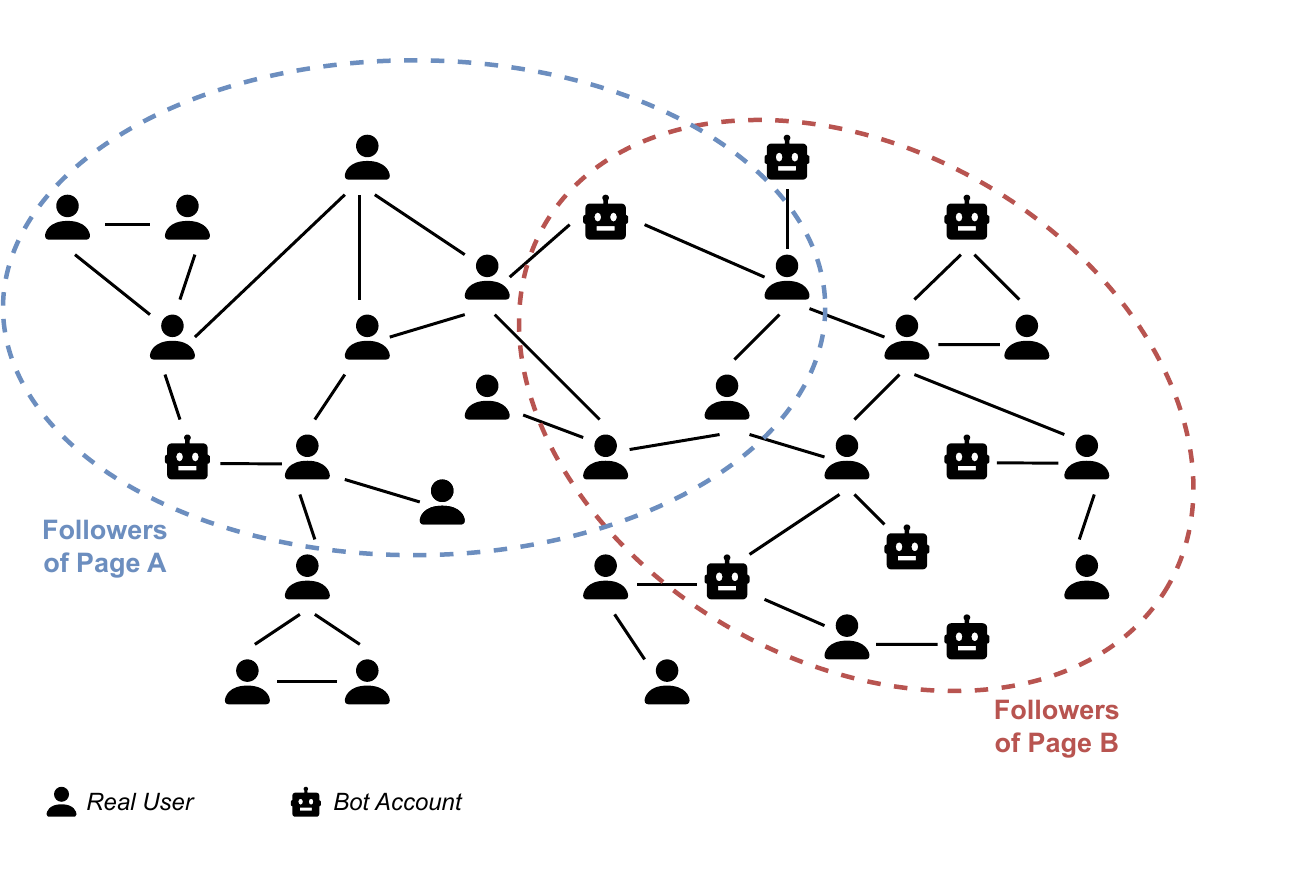}
 \caption{Quantifying bots among the followers of a social network page. An anomalous prevalence of bots may be an indication that Page B has been artificially constructed to be an echo chamber for the spread of misinformation.}
 \label{fig:bots-scenario}
\end{figure}

Social networks have become central to modern communication, shaping public discourse and influencing societal trends.
However, they also present significant challenges, including the spread of misinformation \citep{DelVicario2016}, the formation of echo chambers \citep{Cinelli2021}, and the manipulation of online narratives through automated bots \citep{Stella2019}.
The proliferation of echo chambers and automated bots on social media has significantly altered the landscape of information dissemination, leading to widespread misinformation and polarization.
Echo chambers arise when individuals interact primarily within networks that reinforce their pre-existing beliefs, limiting exposure to diverse perspectives and reducing critical engagement with opposing viewpoints.
Social media algorithms exacerbate this phenomenon by curating content that aligns with user preferences, ultimately reinforcing biases and facilitating the spread of misinformation. Simultaneously, the deployment of bots --- automated accounts programmed to manipulate online discourse --- amplifies deceptive narratives by artificially boosting engagement and creating false consensus.
These bots can be strategically employed to distort public opinion, influence elections, and undermine trust in credible sources \citep{Stella2019}.
The combined effects of these mechanisms contribute to a fragmented and distorted information ecosystem, weakening democratic processes and informed decision-making.
Such tactics may be employed as a concerted effort by a malicious actor, posing a serious security threat \citep{Hellman2024}.
In Figure~\ref{fig:bots-scenario} we present an example of a social network containing real users and bot accounts.
Both types of accounts may interact and form relationships with each other.
In addition, accounts may follow certain social pages that publish posts and other content that users can comment on and share with their contacts.
Quantifying the fraction of bot accounts that subscribe to a social page can be an indication of whether the interactions are artificially inflated to construct an echo chamber by a malicious actor. In our example, Page~B presents an unusual fraction of bot accounts compared to other social pages such as Page~A.
XNQ can enable social media to efficiently identify such pages or user groups, resulting in increased effectiveness in content moderation.
This application scenario represents a slight variation of the tasks of Section~\ref{sec:experiments}, which mostly have dealt with quantifying classes of users in social networks.

% -----------------------------------------------------------

\paragraph{Network quantification in the social sciences and  political science}

\begin{figure}[t]
 \includegraphics[width=\textwidth]{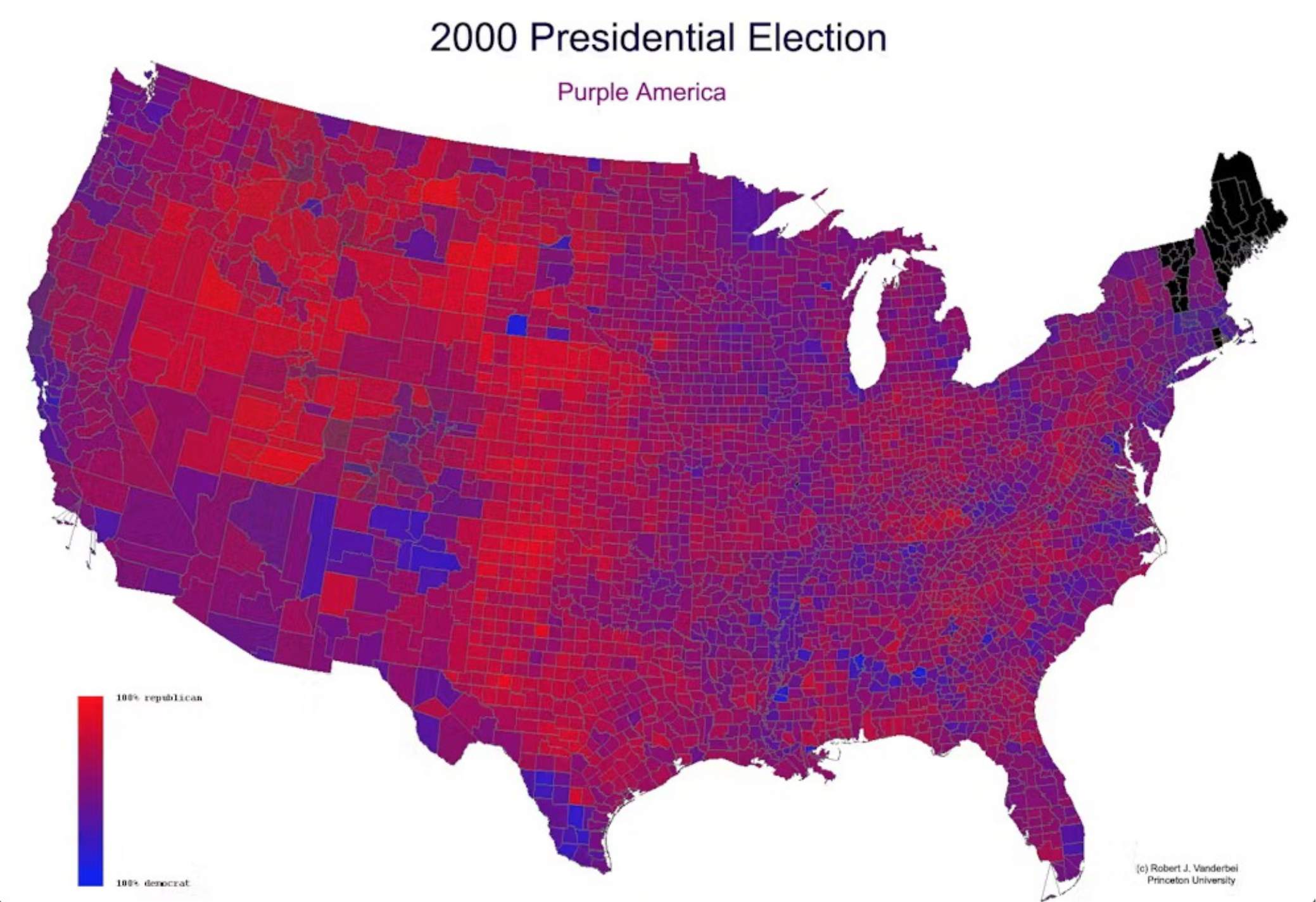}
 \caption{2000 Presidential Election results in the US, at county level. (Image from \url{https://tinyurl.com/2nxy8rzw}, \copyright\ Robert Vanderbei, Princeton University; reprinted with permission of the author.)}
 
 \label{fig:US}
\end{figure}

As recalled in the introduction, quantification finds many applications in disciplines, such as political science, the social sciences, and epidemiology, that have an inherent focus on populations, rather than on individuals. 
For instance, in political science, a researcher may want to predict, for an upcoming election, the prevalence of votes for the different parties (say, Democratic vs.\ Republican) among the population of a certain nation, and how these prevalence values vary geographically (e.g., across different municipalities, provinces, and regions of the country of interest). Here, individuals belonging to the population may each be represented by a set of covariates assumed to predict the phenomenon under study. 
The above inference can straightforwardly be obtained via quantification at the desired level of granularity, e.g., quantifying on a specific province or on a specific municipality of that province.
An interesting observation is that (see Figure~\ref{fig:US} for an example)
the distribution of electoral votes tends to vary \textit{smoothly} in space. In other words, the distributions that characterize two neighbouring municipalities are unlikely to be radically different, and their difference is likely to be smaller than the difference between two faraway municipalities (the same holds at province level, region level, etc.). This suggests that NQ may be more effective than standard quantification, once geographic proximity (i.e., the relation ``$x$ and $y$ are neighbouring geographical units'') is represented as an undirected graph. (In this case, the resulting graph is likely characterized by a high degree of homophily.)

% -----------------------------------------------------------

\paragraph{Network quantification in epidemiology}
A similar case can be made for the geographic distribution of many other phenomena, be they of interest for political science, the social sciences, epidemiology, etc. For example, when studying the spread of infectious diseases (e.g., HIV, or COVID-19), both social and geographical proximity are core factors \citep{Rothenberg2005,Kuchler2022}.
In this scenario, estimating the fraction of infected people within a certain geographical area may be achieved via quantification, and NQ techniques may deliver more accurate results than standard quantification techniques by leveraging, as in the case of election result prediction, proximity information. As noted above, here the notion of ``proximity'' may be different from the previous case; e.g., two towns may be considered ``neighbouring geographical units'' not only if they lie next to each other, but also if a high-frequency flight connects them.
Estimating the prevalence of infections in different geographic areas or within specific groups is extremely useful for planning an effective response to the health crisis and for the informed decision-making process concerning health policies \citep{Madhav2017}.
An accurate quantification of the infected populations can improve: the resource allocation of medical supplies and personnel; the containment measures, such as lockdowns and travel restrictions; and the planning of vaccination campaigns.

% -----------------------------------------------------------

\section{Conclusions}\label{sec:conclusions}
\noindent We have presented XNQ, a novel model tailored to the many
challenges of NQ, which integrates randomized recursive graph neural
networks, a customized calibrated readout for quantification, and a
downstream powerful quantifier based on the EM algorithm. Our
extensive evaluation shows that XNQ improves at this task by
effectively exploiting the graph structure of the data while scaling
seamlessly to hundreds of thousands of nodes without being impaired by
common issues of large-scale networks such as heterophily. These
results place XNQ at the forefront of NQ research and pave the way for
its application to further real-world case studies. In future
research, we plan to extend our approach to the multi-class case, and
to investigate \emph{non-aggregative} NQ methods, i.e., methods that
estimate class priors without assigning labels to (or computing
posterior probabilities for) individual nodes. Unlike the aggregative
methods, non-aggregative ones have the additional advantage that no
inference at the individual level is performed; this is desirable in
applications such as measuring the fairness (i.e., absence of bias) of
a model with respect to sensitive attributes~\citep{Fabris:2023pj}.

% -----------------------------------------------------------

\section*{Declarations}

\bmhead{Funding} This work was partially supported by: Project
DEEP-GRAPH, funded by the Italian Ministry of University and Research
(MUR) PRIN 2022 (project code: 2022YLRBTT, CUP: I53C24002440006);
project PNRR, PE00000013, ``FAIR - Future Artificial Intelligence
Research'', Spoke 1, funded by European Commission under
NextGeneration EU programme (CUP B53D22000980006); project
``Quantification under Dataset Shift'' (QuaDaSh), funded by the
European Union under the NextGenerationEU funding scheme (CUP
B53D23026250001); Project PAN-HUB, funded by the Italian Ministry of Health (POS 2014--2020, project ID: T4-AN-07, CUP: I53C22001300001).

\bmhead{Conflicts of interest/Competing interests} The authors have no
relevant financial or non-financial interests to disclose.

\bmhead{Ethics approval and consent to participate} Not applicable.

\bmhead{Consent for publication} Not applicable.

\bmhead{Data availability} Data is publicly available from referred
sources.

\bmhead{Code availability} The code will be publicly released upon
acceptance. For reviewing purposes only, an anonymized version is
available at
\url{https://github.com/marcopodda/network-quantification}.

\bmhead{Author contributions} A.Micheli and F.S. contributed to the study
conception and design. Material preparation, data collection and
analysis were performed by W.S., D.T., A.Moreo, and M.P. The initial
manuscript was written by W.S., using feedback from all the other
authors. All authors have contributed to the revision of the initial
manuscript, and have read and approved the final version.

% -----------------------------------------------------------

\begin{appendices}

 % -----------------------------------------------------------

 \section{Quantification methods}
 \label{app:quantification-methods}
 \noindent Given a hard classifier $\mathfrak{h}$ and a sample set
 $\sample$, \textit{Classify and Count (CC)} is defined as:
 \begin{align}
 \begin{split}
 \label{eq:CC}
 \hat{p}_{\sample}^{\mathrm{CC}}(\oplus) =
 \frac{1}{|\sample|}\sum_{\boldsymbol{x}_v\in \sample}
 \mathds{1} \left[ \mathfrak{h}(\boldsymbol{x}_v)=\oplus \right]
 \end{split}
 \end{align} 
 i.e., the prevalence of class $\oplus$ is estimated as the number of
 times it is predicted by $\mathfrak{h}$ in $\sample$ divided by
 the number of samples in $\sample$. \textit{Adjusted Classify
 and Count (ACC)} attempts to correct the estimates returned by
 CC. It is defined as: 
 \begin{align}
 \label{eq:acc4}
 \hat{p}_{\sample}^{\operatorname{ACC}}(\oplus) =
 \frac{\hat{p}_{\sample}^{\operatorname{CC}}(\oplus)
 -\hat{\fpr}_{\mathfrak{h}}}{\hat{\tpr}_{\mathfrak{h}} 
 - \hat{\fpr}_{\mathfrak{h}}}
 \end{align}
 where $\hat{\tpr}_{\mathfrak{h}}$ and $\hat{\fpr}_{\mathfrak{h}}$
 are estimates of the true positive and false positive rates obtained
 by hold-out or $k$-fold cross-validation on the training set
 $\mathcal{V}_\text{labelled}$. \textit{Probabilistic Classify and
 Count (PCC)} is a probabilistic counterpart of CC, which replaces
 the hard estimates with expected counts computed from the posterior
 probabilities of a calibrated probabilistic classifier
 $\mathfrak{s}$:
 \begin{align}
 \label{eq:pcc}
 \hat{p}_{\sample}^{\mathrm{PCC}}(\oplus) & = 
 \frac{1}{|\sample|}\sum_{\boldsymbol{x}_v\in 
 \sample}\Pr(Y=\oplus|X=\boldsymbol{x}_v) = 
 \frac{1}{|\sample|}\sum_{\boldsymbol{x}_v\in 
 \sample} \mathfrak{s}(\boldsymbol{x}_v)
 \end{align}
 Similarly, \textit{Probabilistic Adjusted Classify and Count (PACC)}
 is a probabilistic counterpart of ACC where the CC estimate is
 replaced with the PCC estimate, with $\hat{\tpr}_{\mathfrak{s}}$ and
 $\hat{\fpr}_{\mathfrak{s}}$ as probabilistic counterparts of
 $\hat{\tpr}_{\mathfrak{h}}$ and $\hat{\fpr}_{\mathfrak{h}}$,
 respectively:
 \begin{align}
 \label{eq:pacc}
 \hat{p}_{\sample}^{\operatorname{PACC}}(\oplus)=
 \frac{\hat{p}_{\sample}^{\operatorname{PCC}}(\oplus)
 -\hat{\fpr}_{\mathfrak{s}}}{\hat{\tpr}_{\mathfrak{s}} -
 \hat{\fpr}_{\mathfrak{s}}}
 \end{align}
 \textit{HDy} is a probabilistic binary quantification method that
 views quantification as the problem of minimizing the divergence
 (measured in terms of the Hellinger Distance) between two
 distributions of posterior probabilities returned by a soft
 classifier $\mathfrak{s}$, one coming from the unlabelled examples
 and the other coming from a validation set. 
 HDy relies on histograms for modelling the class-conditional distributions of the validation posteriors coming from positive instances ($H_{\oplus}$), and those coming from negative instances ($H_{\ominus}$).
 HDy then looks for the
 mixture parameter $\alpha$ that yields the best match between the validation
distribution (the mixture $\alpha H_{\oplus} + (1-\alpha) H_{\ominus}$) and the test distribution (a histogram of the test posteriors), returning $\alpha$ as
 the estimated prevalence for class $\oplus$. The \textit{DyS} method
 is basically HDy with the Topsøe distance used in place of the
 Hellinger distance.

 % -----------------------------------------------------------

 \section{Convolutional Graph Neural Networks}\label{app:conv-gnn}
 \noindent In the ablation experiments, we use different
 convolutional GNNs to study their performances in comparison with
 XNQ. In Table~\ref{tab:conv}, we briefly describe them for
 completeness, using the notation developed in
 Section~\ref{sec:background}. We use $1 \leq \ell \leq L$ to
 indicate the number of convolutional layers,
 $\boldsymbol{h}_v^{(\ell)}$ to denote the embedding of node $v$
 computed at the $\ell^{\mathrm{th}}$ layer, and
 $\boldsymbol{W}^{(\ell)}$ as the $\ell^{\mathrm{th}}$ matrix of
 learnable weights specific for each layer. In the table,
 $\tilde{\mathcal{N}}(v) = \mathcal{N}(v)\cup \{v\}$ is the closed
 neighbourhood of $v$, $\alpha_{ij}$ are attention coefficients
 computed by comparing pairs of neighbouring embeddings, $\epsilon$ is
 a small learnable constant and $\mathrm{MLP}$ is a multilayer
 perceptron.
 \begin{table}[h!]
 \caption{Message passing variants used in the ablation studies.}
 \label{tab:conv}
 \centering
 \begin{tabular}{ll}
 \toprule
 \textbf{Message Passing}& $\boldsymbol{h}_v^{(\ell)}$ \\
 \midrule
 GCN & 
 $\mathrm{sigmoid}\left(\boldsymbol{W}^{(\ell)} 
 \sum_{u \in \tilde{\mathcal{N}}(v)} 
 \frac{1}{|\tilde{\mathcal{N}}(v)|}\boldsymbol{h}_u^{(\ell
 1)}\right)$ \\
 GAT & 
 $\text{ReLU}\left( 
 \alpha_{vv}\boldsymbol{W}^{(\ell)}\boldsymbol{h}_v^{(\ell
 -1)} 
 + \sum_{u \in \mathcal{N}(v)} 
 \alpha_{vu}\boldsymbol{W}^{(\ell)}\boldsymbol{h}_u^{(\ell
 -1)}\right)$\\
 GIN & 
 $\mathrm{MLP}\left((1+\epsilon)\, 
 \boldsymbol{h}_v^{(\ell-1)} + \sum_{u \in \mathcal{N}(v)} 
 \boldsymbol{h}_v^{(\ell-1)}\right)$\\
 \bottomrule
 \end{tabular}
 \end{table}
 
% -----------------------------------------------------------

 \section{Statistical significance}\label{app:statistical-significance}
In order to assess whether the differences in performance between our proposed XNQ and the alternative methods are statistically significant, we compare the different models to XNQ across datasets using Bayesian analysis, following the work of \cite{benavoli2017}.

The test first computes the posterior distribution of the differences in performance between the two models on multiple datasets. Then, it draws samples from the posterior distribution to estimate the number of times one model outperforms the other, or how many times they perform equivalently (where equivalence is established if performances fall in a predefined region around the posterior mean, called \textit{rope}). Here, we set the rope to the AE/RAE standard deviation of XNQ across the different datasets. 

The results are presented in Table~\ref{tab:comparison}. From the table, it is possible to see that, irrespective of the model compared to XNQ, the posterior probability that the alternative method would perform better than XNQ is always 0 (except the XNQ-ENQ comparison with RAE on the binary datasets, where however the probability that ENQ is better than XNQ is only of 3.2\%). There are some situations where ties are likely, though: for example, if the downstream quantifier is ACC rather than SLD, there is approximately a 50\% chance of a performance tie. However, we argue in favor of SLD as our quantifier of choice, which is able to achieve top-tier performance with solid theoretical guarantees (in particular, Fisher consistency).
\begin{table}[]
\caption{Bayesian test comparison of XNQ's performances versus the different baselines and ablation models used in this study. Each row expresses the posterior probability that XNQ performs better than the competitor. Highest probability is \underline{underlined}. Results using RAE as main metric are shown within parentheses.}\label{tab:comparison}
 \centering
\begin{tabular}{lccc}
\toprule
\textbf{XNQ vs.} & \textbf{$p$(XNQ is worse)} & \textbf{$p$(XNQ is equal)} & \textbf{$p$(XNQ is better)} \\ 
\midrule
\multicolumn{4}{l}{Baselines on binary datasets (Tables \ref{tab:results}-\ref{tab:results-rae})} \\ 
\midrule
wvRN & 0.000 (0.000) & 0.010 (0.001) & \underline{0.990} (\underline{0.999}) \\
CDQ & 0.000 (0.000) & 0.003 (0.001) & \underline{0.997} (\underline{0.999}) \\
ENQ & 0.000 (0.032) & 0.001 (0.001) & \underline{0.999} (\underline{0.967}) \\
\midrule
\multicolumn{4}{l}{Baselines on multi-class datasets (Tables \ref{tab:results-multi}-\ref{tab:results-multi-rae})} \\ 
\midrule
wvRN & 0.000 (0.000) & 0.395 (0.009) & \underline{0.605} (\underline{0.991}) \\
CDQ & 0.000 (0.000) & 0.092 (0.143) & \underline{0.908} (\underline{0.857}) \\
ENQ & 0.000 (0.000) & \underline{0.844} (0.038) & 0.156 (\underline{0.962}) \\
\midrule
\multicolumn{4}{l}{Node embedder ablation (Table \ref{tab:ablation})} \\
\midrule
LR & 0.000 & 0.259 & \underline{0.741} \\
GCN & 0.000 & 0.015 & \underline{0.985} \\
GAT & 0.000 & 0.009 & \underline{0.991} \\
GIN & 0.000 & 0.003 & \underline{0.997} \\
\midrule
\multicolumn{4}{l}{Quantifier ablation (Table \ref{tab:ablation})} \\
\midrule
CC & 0.000 & 0.001 & \underline{0.999} \\
ACC & 0.000 & \underline{0.541} & 0.449 \\
PCC & 0.000 & 0.001 & \underline{0.999} \\
PACC & 0.000 & \underline{0.789} & 0.211 \\
HDy & 0.000 & 0.141 & \underline{0.859} \\
DyS & 0.000 & \underline{0.789} & 0.211 \\ 
\bottomrule
\end{tabular}
\end{table}

% -----------------------------------------------------------

\end{appendices}

\bibliography{bibliography}

% -------------------------------------------------------------------------

%% -------------------------------------------------------

% \newpage

% \appendix

% \pagenumbering{roman}
% \setcounter{page}{1}

\end{document}